\documentclass{article} % For LaTeX2e
\usepackage{macros}
\usepackage{xr}
\usepackage{epstopdf}
\title{Risk-Sensitive and Robust Decision-Making:\\  a CVaR Optimization Approach}
\author{
Yinlam Chow\thanks{Institute of Computational \& Mathematical Engineering, Stanford University. Stanford CA, USA 94305},\,\, Aviv Tamar\thanks{Electrical Engineering Department, The Technion - Israel Institute of Technology, Haifa, Israel 32000}, \,\, Shie Mannor\thanks{Electrical Engineering Department, The Technion - Israel Institute of Technology, Haifa, Israel 32000},\,\, Marco Pavone\thanks{Aeronautics and Astronautics Department, Stanford University, Stanford CA, USA 94305} }%\\

\newcommand{\BEAS}{\begin{eqnarray*}}
\newcommand{\EEAS}{\end{eqnarray*}}
\newcommand{\BEQ}{\begin{equation}}
\newcommand{\EEQ}{\end{equation}}
\newcommand{\BIT}{\begin{itemize}}
\newcommand{\EIT}{\end{itemize}}

\usepackage{color}
\usepackage{amsmath}
\usepackage{amssymb}
\usepackage{graphicx}
\usepackage{comment,xspace}
\usepackage{fancybox}
\newcommand{\real}{{\mathbb{R}}}
\newcommand{\reals}{\real}

\renewcommand{\natural}{{\mathbb{N}}}

\newtheorem{proposition}[theorem]{Proposition}

\newtheorem{assumption}{Assumption}

\newcommand{\argmin}{\operatornamewithlimits{argmin}}

\def\R{R}

\def\V{V}

\def\X{\mathcal{X}}

\newcommand{\pr}{P} % probability measure
\newcommand{\mdp}{\mathcal{M}} % policy parameter
\newcommand{\St}{{\mathcal{X}}} % state space
\newcommand{\Ac}{{\mathcal{A}}} % action space
\newcommand{\Exp}[1]{\mathbb{E}\left[ #1 \right]}
\newcommand{\ExpW}[2]{\mathbb{E}_{#2} \left[ {#1} \right]}
\newcommand{\U}{{\mathcal{U}}} % coherent risk set U

\newcommand{\alggiven}{\textbf{Given:}}

\newcommand{\interpY}{\mathbf Y} %interpolation set
\newcommand{\interpI}{\mathcal I} %interpolation function
\newcommand{\bellint}{\mathbf T_{ \interpI}} %interpolation function
\newcommand{\mnorm}[1]{{\left\vert\kern-0.25ex\left\vert\kern-0.25ex\left\vert #1
    \right\vert\kern-0.25ex\right\vert\kern-0.25ex\right\vert}}%{||| {#1} |||} 

\begin{document}

\maketitle

%%%%%%%%%%%%%%%%%%%%%%%%%%%%%%%%%%%%%%%%%%%%%%%%%%%%%%%%%%%%%%
%%%%%%%%%%%%%%%%%%%%%%%%%%%%%%%%%%%%%%%%%%%%%%%%%%%%%%%%%%%%%%
%%%%%%%%%%%%%%%%%%%%%%%%%%%%%%%%%%%%%%%%%%%%%%%%%%%%%%%%%%%%%%
%%%%%%%%%%%%%%%%%%%%%%%%%%%%%%%%%%%%%%%%%%%%%%%%%%%%%%%%%%%%%%
%%%%%%%%%%%%%%%%%%%%%%%%%%%%%%%%%%%%%%%%%%%%%%%%%%%%%%%%%%%%%%

\vspace{-0.25in}
\begin{abstract}
In this paper we address the problem of decision making within a Markov decision process (MDP) framework where risk and modeling errors are taken into account. Our approach is to  minimize a risk-sensitive conditional-value-at-risk (CVaR) objective, as opposed to a standard risk-neutral expectation. We refer to such problem as CVaR MDP. Our first contribution is to show that a CVaR objective, besides capturing risk sensitivity, has an alternative interpretation as expected cost under worst-case modeling errors, for a given error budget. This result, which is of independent interest,  motivates CVaR MDPs as a unifying framework for risk-sensitive and robust decision making. Our second contribution is to present an approximate  value-iteration algorithm for CVaR MDPs and analyze its convergence rate. To our knowledge, this is the first solution algorithm for CVaR MDPs that enjoys error guarantees. Finally, we present results from numerical experiments that corroborate our theoretical findings and show the practicality of our approach.
\end{abstract}

%%%%%%%%%%%%%%%%%%%%%%%%%%%%%%%%%%%%%%%%%%%%%%%%%%%%%%%%%%%%%%%%%%%%%%%%%%%%%%%%
\vspace{-0.1in}
\section{Introduction}
\vspace{-0.1in}
Decision making within the Markov decision process (MDP) framework typically involves the minimization of a risk-neutral performance objective, namely the \emph{expected} total discounted cost \cite{Ber2012DynamicProgramming}. This approach, while very popular, natural, and attractive from a computational standpoint, neither takes into account the \emph{variability} of the cost (i.e., fluctuations around the mean), nor its \emph{sensitivity} to modeling errors, which may significantly affect overall performance \cite{mannor2007bias}. Risk-sensitive MDPs \cite{Howard1972Risk} address the first aspect by replacing the risk-neutral expectation with a \emph{risk-measure} of the total discounted cost, such as variance, Value-at-Risk (VaR), or Conditional-VaR (CVaR). Robust MDPs \cite{nilim_robust_2005}, on the other hand, address the second aspect by defining a set of plausible MDP parameters, and optimize decision with respect to the worst-case scenario.

In this work we consider risk-sensitive MDPs with a CVaR objective, referred to as CVaR MDPs. CVaR \cite{artzner1999coherent, rockafellar2000optimization} is a risk-measure that is rapidly gaining popularity in various engineering applications, e.g., finance, due to its favorable computational properties \cite{artzner1999coherent} and superior ability to safeguard a decision maker from the ``outcomes that hurt the most" \cite{serraino2013conditional}. In this paper,  by \emph{relating risk to robustness}, we derive a novel result that further motivates the usage of a CVaR objective in a decision-making context. Specifically, we show that the CVaR of a discounted cost in an MDP is \emph{equivalent} to the expected value of the same discounted cost in presence of worst-case perturbations of the MDP parameters (specifically, transition probabilities), provided that such perturbations are within a certain error budget. This result suggests CVaR MDP as a method for decision making under \emph{both} cost variability \emph{and} model uncertainty, motivating it as \emph{unified framework} for planning under uncertainty. 

%We guarantee that our approach is computationally tractable, and demonstrate its efficacy on a grid-world navigation domain with obstacles.

\vspace{-0.05in}
\emph{Literature review}: Risk-sensitive MDPs have been studied for over four decades, with earlier efforts focusing on exponential utility \cite{Howard1972Risk}, mean-variance \cite{sobel_variance_1982}, and percentile risk criteria \cite{filar_percentile_1995} . Recently, for the reasons explained above, several authors have investigated CVaR MDPs \cite{,rockafellar2000optimization}. Specifically, in \cite{borkar2014risk}, the authors propose a dynamic programming algorithm for finite-horizon risk-constrained MDPs where risk is measured according to CVaR. The algorithm is proven to asymptotically converge to an optimal risk-constrained policy. However, the algorithm involves computing  integrals over continuous variables (Algorithm 1 in \cite{borkar2014risk}) and, in general, its implementation appears particularly difficult. In \cite{bauerle2011markov}, the authors investigate the structure of CVaR optimal policies and show that a Markov policy is optimal on an augmented state space, where the additional (continuous) state variable is represented  by the running cost. In   \cite{haskell2014convex}, the authors leverage such result to design an algorithm for CVaR MDPs that relies on discretizing occupation measures in the augmented-state MDP. This approach, however, involves solving a non-convex program via a sequence of linear-programming approximations, which can only shown to converge asymptotically. A different approach is taken by \cite{chow2014cvar} and \cite{tamar2015optimizing}, which consider a finite dimensional parameterization of control  policies, and show that a CVaR MDP can be optimized to a \emph{local} optimum using stochastic gradient descent (policy gradient). A recent result by Pflug and Pichler \cite{pflug2012time} showed that CVaR MDPs admit a dynamic programming formulation by using a state-augmentation procedure different from the one in \cite{bauerle2011markov}. The augmented state  is also continuous, making the design of a solution algorithm challenging. 

\vspace{-0.05in}
\emph{Contributions}: The contribution of this paper is twofold. First, as discussed above, we provide a novel interpretation for CVaR MDPs in terms of robustness to modeling errors. This result is of independent interest and further motivates the usage of CVaR MDPs for decision making under uncertainty. Second, we provide a new optimization algorithm for CVaR MDPs, which leverages the state augmentation procedure introduced by Pflug and Pichler \cite{pflug2012time}. We overcome the aforementioned computational challenges (due to the continuous augmented state) by designing an algorithm that  merges approximate value iteration \cite{Ber2012DynamicProgramming} with linear interpolation. Remarkably, we are able to provide explicit error bounds and convergence rates based on contraction-style arguments. In comparison to the algorithms in \cite{borkar2014risk,haskell2014convex,chow2014cvar,tamar2015optimizing}, our approach leads to finite-time error guarantees, with respect to the \emph{globally} optimal policy. In addition, our algorithm is significantly simpler than previous methods, and calculates the optimal policy \emph{for all} CVaR confidence intervals and initial states simultaneously. The practicality of our approach is  demonstrated in numerical experiments involving planning a path on a grid with thousand of states. To the best of our knowledge, this is the first algorithm to compute globally-optimal policies for non-trivial CVaR MDPs.

%It is important to distinguish between the CVaR of the total discounted return, as explored in this work, and the dynamic (a.k.a. iterated) CVaR, as studied in \cite{ruszczynski2010risk,osogami2012robustness,petrik2012approximate}. Simply stated, the dynamic CVaR assumes the worst \emph{at each} time step, and consequentially results in very conservative policies.

%CVaR MDPs have been recently considered by several authors \cite{borkar2014risk,haskell2014convex,chow2014cvar,tamar2015optimizing}. While the algorithms in these works are guaranteed to converge asymptotically, no convergence rates nor error guarantees are provided. In this work we provide a new CVaR optimization algorithm, based on a recent result by Pflug and Pichler \cite{pflug2012time}. Our algorithm resembles approximate value iteration \cite{Ber2012DynamicProgramming} on a state-space augmented by a continuous auxiliary variable, and enables us to provide explicit error bounds and convergence rates based on contraction-style arguments.
%
%

%\paragraph{Paper Structure}
\vspace{-0.05in}
\emph{Organization}:  This paper is structured as follows. In Section \ref{sec:prelim} we provide background on CVaR and MDPs, we state the problem we wish to solve (i.e., CVaR MDPs), and motivate  the CVaR MDP formulation by establishing a novel  relation between CVaR and model perturbations. Section \ref{subsec:Risk-Bellman} provides the basis for our solution algorithm, based on a Bellman-style equation for the CVaR. Then, in Section \ref{sec:algorithm_interpolation} we present our algorithm and correctness analysis. in Section \ref{sec:experiments} we evaluate our approach via numerical experiments. Finally, in Section \ref{sec:conclusion}, we draw some conclusions and discuss directions for future work.

%%%%%%%%%%%%%%%%%%%%%%%%%%%%%%%%%%%%%%%%%%%%%%%%%%%%%%%%%%%%%%%%%%%%%%%%%%%%%%%%
\vspace{-0.1in}
\section{Preliminaries, Problem Formulation, and Motivation}\label{sec:prelim}
\vspace{-0.1in}
\subsection{Conditional Value-at-Risk}
\label{subsec:coherent}
\vspace{-0.1in}
Let $Z$ be a bounded-mean random variable, i.e.,~$\E[|Z|]<\infty$, on a probability space $(\Omega, \mathcal F,\mathbb P)$, with cumulative distribution function $F(z)=\mathbb{P}(Z\leq z)$. In this paper we interpret $Z$ as a cost. The {\em value-at-risk} (VaR) at confidence level $\alpha\in (0,1)$ is the $1-\alpha$ quantile of $Z$, i.e., VaR$_\alpha(Z) = \min\big\{z\mid F(z)\geq\alpha\big\}$.
%
%\vspace{-0.1in}
%\begin{small}
%\begin{equation}
%\label{eq:VaR}
%\text{VaR}_\alpha(Z) = \min\big\{z\mid F(z)\geq\alpha\big\}.
%\end{equation}
%\end{small}
%\vspace{-0.15in}
%
%The minimum in~\eqref{eq:VaR} is attained because $F$ is non-decreasing and right-continuous in $z$. When $F$ is continuous and strictly increasing, VaR$_\alpha(Z)$ is the unique $z$ satisfying $F(z)=\alpha$, otherwise, \eqref{eq:VaR} can have no solution or a whole range of solutions. Although VaR is a popular risk measure, it suffers from being unstable and difficult to work with numerically when $Z$ is not normally distributed, which is often the case as loss distributions tend to exhibit fat tails or empirical discreteness. Moreover, VaR is not a {\em coherent} risk measure~\cite{Artzner99CM} and more importantly does not quantify the losses that might be suffered beyond its value at the $\alpha$-tail of the distribution~\cite{Rockafellar00OC}. An alternative measure that addresses most of the VaR's shortcomings is {\em conditional value-at-risk}, CVAR$_\alpha(Z)$, which is the mean of the $\alpha$-tail distribution of $Z$. If there is no probability atom at VaR$_\alpha(Z)$, CVaR$_\alpha(Z)$ has a unique value that is defined as
%
%Here the minimum is attained because $F$ is non-decreasing and right-continuous in $z$.
%When $F$ is continuous and strictly increasing, VaR$_\alpha(Z)$ is the unique $z$ satisfying $F(z)=\alpha$, otherwise, the VaR equation can have no solution or a whole range of solutions.
The \emph{conditional value-at-risk} (CVaR) at confidence level $\alpha\in (0,1)$ is defined as~\cite{rockafellar2000optimization}:
\begin{equation}
\label{eq:CVaR2}
\text{CVaR}_\alpha(Z) = \min_{w\in\reals}\Big\{w + \frac{1}{\alpha}\mathbb E\big[(Z-w)^+\big]\Big\},
\end{equation}
%\vspace{-0.1in}
where $(x)^+=\max(x,0)$ represents the positive part of $x$.
If there is no probability atom at VaR$_\alpha(Z)$, it is well known that CVaR$_\alpha(Z) = \E\big[Z\mid Z\geq \text{VaR}_\alpha(Z)\big]$. Therefore, CVaR$_\alpha(Z)$ may be interpreted as the worst case expected value of $Z$, conditioned on the $\alpha$-portion of the tail distribution.
It is well known that CVaR$_\alpha(Z)$ is decreasing in $\alpha$, $\text{CVaR}_1(Z)$ equals to $\mathbb E(Z)$, and $\text{CVaR}_\alpha(Z)$ tends to $\max(Z)$ as $\alpha\downarrow 0$. During the last decade, the CVaR risk-measure has gained popularity in financial applications, among others. It is especially useful for controlling rare, but potentially disastrous events, which occur below the $1-\alpha$ quantile, and are neglected by the VaR \cite{serraino2013conditional}. Furthermore, CVaR enjoys desirable axiomatic properties, such as coherence \cite{artzner1999coherent}. We refer to \cite{uryasev2010var} for further motivation about CVaR and a comparison with other risk measures such as VaR.

%Equipped with this risk modeling property, using CVaR risk metric in optimization problems has numerous financial and engineering applications, especially when the probability of undesirable events is small but the loss incurred can still be enormous.
%
%Although VaR is a popular risk measure, it suffers from being unstable and difficult to work with numerically when $Z$ is not normally distributed, which is often the case as loss distributions tend to exhibit fat tails or empirical discreteness. Moreover, VaR is not a {\em coherent} risk measure~\cite{artzner1999coherent} and more importantly does not quantify the losses that might be suffered beyond its value at the $\alpha$-tail of the distribution~\cite{rockafellar2000optimization}. An alternative measure that addresses most of the VaR's shortcomings is {\em conditional value-at-risk}, CVaR$_\alpha(Z)$, which is the mean of the $\alpha$-tail distribution of $Z$. \text{CVaR} belongs to the well-known class of \emph{coherent} risk measures \cite{artzner1999coherent} for which it is \emph{convex}, \emph{translational invariant}, \emph{positive homogenous} and \emph{monotonic}.

A useful property of CVaR, which we exploit in this paper, is its alternative dual representation \cite{artzner1999coherent}:
\begin{equation}\label{eq:coherent_cvar}
\text{CVaR}_\alpha(Z)=\max_{\xi\in \U_{\text{CVaR}}(\alpha,\mathbb P)} \mathbb E_{\xi}[ Z],
\end{equation}
where $\mathbb E_{\xi}[ Z]$ denotes the $\xi$-weighted expectation of $Z$, and the \emph{risk envelop} $\U_{\text{CVaR}}$ is given by
$
\U_{\text{CVaR}}(\alpha,\mathbb P)=\left\{\xi:\xi(\omega)\in\bigg[0,\frac{1}{\alpha}\bigg],\,\int_{\omega\in\Omega} \xi(\omega)\mathbb P(\omega) d\omega=1\right\}.
$
%\vspace{-0.1in}
Thus, the CVaR of a random variable $Z$ may be interpreted as the worst-case expectation of $Z$, under a perturbed distribution $\xi \mathbb P$.

In this paper, we are interested in the CVaR of the total discounted cost in a sequential decision-making setting, as discussed next.

\subsection{Markov Decision Processes}
An MDP is a tuple $\mdp=(\St,\Ac,C,P,x_0,\gamma)$, where $\St$ and $\Ac$ are finite state and action spaces; $C(x,a)\in[-C_{\max},C_{\max}]$ is a bounded deterministic cost;
$P(\cdot|x,a)$ is the transition probability distribution; $\gamma\in[0,1)$ is the discounting factor, and $x_0$ is the initial state. (Our results easily generalize to random initial states and random costs.)

Let the space of admissible histories up to time $t$ be $H_t = H_{t-1} \times \mathcal X$, for $t\geq 1$, and $H_0=\mathcal X$. A generic element $h_t\in H_t$ is of the form $h_t = (x_0,a_0, \ldots , x_{t-1},a_{t-1},x_t)$. Let $\Pi_{H,t}$ be the set of all deterministic history-dependent policies with the property that at each time $t$ the control is a function of $h_t$. In other words, $\Pi_{H,t} := \bigl \{ \mu_0: H_0 \rightarrow \mathcal A,\, \mu_1: H_1 \rightarrow \mathcal A, \ldots,\mu_{t}: H_{t} \rightarrow \mathcal A\} | \mu_j(h_j) \in \mathcal A \text{ for all } h_j\in H_j, \, 1\leq j\leq t \bigr\}$.
We also let $\Pi_{H}=\lim_{t\rightarrow\infty}\Pi_{H,t}$ be the set of all history dependent policies.
\vspace{-0.1in}
\subsection{Problem Formulation}\label{sec:problem}
\vspace{-0.1in}
Let the sequence of random variables $Z_t$ denote the stage-wise costs observed along a state/control trajectory in the MDP model, and let $\mathcal C_{0,T} = \sum_{t=0}^{T}\gamma^t Z_t$ denote the total discounted cost up to time $T$.
The risk-sensitive discounted-cost problem we wish to address is as follows:
\begin{equation}\label{eq:problem}
\min_{\mu\in\Pi_H} \quad \text{CVaR}_\alpha\left(\left. \lim_{T\rightarrow\infty}\mathcal C_{0,T} \right| x_0,\mu\right),
\end{equation}
where $\mu=\{\mu_0,\mu_1,\ldots\}$ is the policy sequence with actions $a_t=\mu_t(h_t)$ for $t\in\{0,1,\ldots\}$. We refer to problem \eqref{eq:problem} as CVaR MDP.  (One may also consider a related formulation combining mean and CVaR, the details of which are presented in the  supplementary material.)
%and $\rho_{\alpha}$ is a spectral risk measure. Except for very limited cases, there is no reason to hope that problem \eqref{eq:problem} should be a tractable problem, since the dependence of the risk measure on $\mu$ may be complex and non-convex.
%In this work, we aim to derive an approximate dynamic programming to search for ``good" policies for \eqref{eq:problem}. In Section \ref{sec:dyn_approx} to \ref{sec:approx_alg}, we first summarize the results when $\rho=\text{CVaR}_\alpha$ and present
%two approximate dynamic programming algorithms.
%Then in Section \ref{sec:SRMs}, we will further generalize this results to SRMs.

The problem formulation in \eqref{eq:problem} directly addresses the aspect of risk sensitivity, as demonstrated by the numerous applications of CVaR optimization in finance (see, e.g., \cite{rockafellar2006master,iyengar2013fast,dowd2007measuring}) and the recent approaches for CVaR optimization in MDPs \cite{borkar2014risk,haskell2014convex,chow2014cvar,tamar2015optimizing}. In the following, we show a new result providing additional motivation for CVaR MDPs, from the point of view of \emph{robustness to modeling errors}.
\vspace{-0.1in}
\subsection{Motivation - Robustness to Modeling Errors}\label{sec:risk_and_robustness}
\vspace{-0.1in}
We show a new result relating the CVaR objective in \eqref{eq:problem} to the worst-case \emph{expected} discounted-cost in presence of worst-case perturbations of the MDP parameters, where the perturbations are budgeted according to the ``number of things that can go wrong." Thus, by minimizing  CVaR, the decision maker also guarantees \emph{robustness} of the policy.

Consider a trajectory $(x_0,\dots,x_T)$ in a finite-horizon MDP problem with transitions $P_t(x_t|x_{t-1})$. We explicitly denote the time index of the transition matrices for reasons that will become clear shortly. The total probability of the trajectory is $\pr{(x_0,\dots,x_T)} = P_0(x_0)P_1(x_1|x_0)\cdots P_T(x_T|x_{T-1})$, and we let $\mathcal C_{0,T}(x_1,\dots,x_T)$ denote its discounted cost, as defined above.

We consider an adversarial setting, where an adversary is allowed to change the transition probabilities at each stage, under some budget constraints. We will show that, for a specific budget and perturbation structure, the expected cost under the worst-case perturbation is equivalent to the CVaR of the cost. Thus, we shall establish that, in this perspective, being risk sensitive is \emph{equivalent} to being robust against model perturbations.

For each stage $1\leq t\leq T$, consider a perturbed transition matrix $\hat{P}_t = P_t \circ \delta_t$, where $\delta_t \in \R^{\St \times \St}$ is a \emph{multiplicative probability perturbation} and $\circ$ is the Hadamard product, under the condition that $\hat{P}_t$ is a stochastic matrix. Let $\Delta_t$ denote the set of perturbation matrices that satisfy this condition, and let $\Delta = \Delta_1 \times \cdots \times \Delta_T$ the set of all possible perturbations to the trajectory distribution.

We now impose a budget constraint on the perturbations as follows. For some budget $\eta \geq 1$, we consider the constraint
\begin{equation}\label{eq:budget_constraint}
    \delta_1(x_1|x_0)\delta_2(x_2|x_1)\cdots \delta_T(x_T|x_{T-1}) \leq \eta, \quad \forall x_1,\dots,x_T\in\St, t=0,\dots,T.
\end{equation}
Essentially, the product in Eq. \eqref{eq:budget_constraint} states that \emph{the worst cannot happen at each time}. Instead, the perturbation budget has to be split (multiplicatively) along the trajectory. We note that Eq. \eqref{eq:budget_constraint} is in fact a constraint on the perturbation matrices, and we denote by $\Delta_\eta \subset \Delta$ the set of perturbations that satisfy this constraint with budget $\eta$.
The following result shows an equivalence between the CVaR and the worst-case expected loss.
\begin{proposition}[Interpretation of CVaR as a Robustness Measure]\label{prop:CVaR_is_robustness}
It holds
\begin{equation}\label{eq:CVaR_is_robustness}
    \text{CVaR}_{\frac{1}{\eta}} \left(\mathcal C_{0,T}(x_1,\dots,x_T)\right) = \sup_{(\delta_1,\dots,\delta_T)\in \Delta_{\eta}} \ExpW{\mathcal C_{0,T}(x_1,\dots,x_T)}{\hat{P}},
\end{equation}
where $\ExpW{\cdot}{\hat{P}}$ denotes expectation with respect to a Markov chain with transitions $\hat{P}_t$.
\end{proposition}

The proof of Proposition \ref{prop:CVaR_is_robustness} is in the supplementary material. It is instructive to compare Proposition \ref{prop:CVaR_is_robustness} with the dual representation of CVaR in  \eqref{eq:coherent_cvar}. Note, in particular, that the perturbation budget in Proposition \ref{prop:CVaR_is_robustness} has a \emph{temporal} structure, which constrains the adversary from choosing the worst perturbation at each time step.

\begin{remark}
An equivalence between robustness and risk-sensitivity was previously suggested by Osogami \cite{osogami2012robustness}. In that study, the \emph{iterated} (dynamic) coherent risk was shown to be equivalent to a robust MDP \cite{iyengar_robust_2005} with a rectangular uncertainty set. The iterated risk (and, correspondingly, the rectangular uncertainty set) is very conservative \cite{xu2006robustness}, in the sense that \emph{the worst can happen at each time step}. In contrast, the perturbations considered here are much less conservative. In general, solving robust MDPs without the rectangularity assumption is NP-hard. Nevertheless, Mannor et. al. \cite{mannor2012lightning} showed that, for cases where the number of perturbations to the parameters along a trajectory is upper bounded (budget-constrained perturbation), the corresponding robust MDP problem is tractable. Analogous to the constraint set (1) in \cite{mannor2012lightning}, the perturbation set in Proposition \ref{prop:CVaR_is_robustness} limits the total number of log-perturbations along a trajectory. Accordingly, we shall later see that optimizing problem \eqref{eq:problem} with perturbation structure \eqref{eq:budget_constraint} is indeed also tractable.
\end{remark}
%\begin{remark}
%\mpmargin{The multiplicative perturbation structure in \eqref{eq:budget_constraint} implies that transitions with zero probability are not perturbed. In practice, when planning for deterministic systems, one may add a small noise to the dynamics. If the noise is small enough, its effect on the expected return would be negligible, while its effect on the CVaR may be significant when $\alpha$ is small. See Section \ref{sec:experiments} for a demonstration.}{Obscure}
%\end{remark}

Next section provides the fundamental theoretical ideas behind our approach to the solution of \eqref{eq:problem}.

\vspace{-0.1in}
\section{Bellman Equation for CVaR}\label{subsec:Risk-Bellman}
\vspace{-0.1in}
In this section, by leveraging a recent result from \cite{pflug2012time}, we present a dynamic programming (DP) formulation for the CVaR MDP problem in \eqref{eq:problem}. As we shall see, the value function in this formulation depends on both the state  and the CVaR confidence level $\alpha$. We then establish important properties of such DP formulation, which will later enable us to derive an efficient DP-based approximate solution algorithm and provide correctness guarantees on the approximation error. All proofs are presented in the supplementary material. 
%However, since $\alpha$ is continuous, actually applying DP is impossible. Nevertheless, we establish important properties of the DP formulation that will later enable us to apply DP using linear interpolation, and provide guarantees on the interpolation error.
%Later we will see that since the augmented state and action spaces are continuous, we need an approximate dynamic programming approach for this formulation, under the framework of approximate value iteration \cite{de2000existence}. To begin with, we first illustrate the dynamic programming results using the CVaR metric.

Our starting point is a recursive decomposition of CVaR, whose proof is detailed in Theorem 10 of \cite{pflug2012time}.
%Assume the control input $a_t$ is induced by policy $\mu_t(h_t)$. Equipped with the MDP formulation and noted that $\mathcal F_t=\{x_0,\ldots,x_t\}$, we provide the following result for a recursive decomposition of CVaR, whose proof is detailed in Theorem 10 of \cite{pflug2012time}.

\begin{theorem}[CVaR Decomposition Theorem, \cite{pflug2012time}]\label{thm:decompose_CVaR}
For any $t\geq 0$, denote by $Z  = (Z_{t+1},Z_{t+2},\dots)$ the cost sequence from time $t+1$ onwards. The conditional CVaR under policy $\mu$, i.e., $\text{CVaR}_{\alpha}(Z\mid H_t,\mu)$, obeys the following decomposition:
\[
\text{CVaR}_{\alpha}(Z\mid H_t,\mu) = \max_{\xi\in \U_{\text{CVaR}}(\alpha, P(\cdot|x_t,a_t))}\mathbb E[\xi(x_{t+1})\cdot \text{CVaR}_{\alpha\xi(x_{t+1})}(Z\mid H_{t+1},\mu)\mid H_t,\mu],
\]
where $a_t$ is the action induced by policy $\mu_t(h_t)$, and the expectation is with respect to~$x_{t+1}$.
\end{theorem}
\vspace{-0.1in}
Theorem \ref{thm:decompose_CVaR} concerns a fixed policy $\mu$; we now extend it to a general DP formulation. Note that in the recursive decomposition  in Theorem \ref{thm:decompose_CVaR} the right-hand side involves CVaR terms with different confidence levels than that in the left-hand side. Accordingly, we augment the state space $\mathcal X$ with an additional continuous state $\mathcal Y=(0,1]$, which  corresponds to the confidence level. For any $x\in \mathcal X$ and $y \in \mathcal Y$, the \emph{value-function} $V(x,y)$ for the augmented state $(x,y)$  is defined as:
\begin{equation*}
    V(x,y)= \min_{\mu\in\Pi_H} \text{CVaR}_y\left(\lim_{T\rightarrow\infty}\mathcal C_{0,T}\mid x_0=x,\mu\right).
\end{equation*}
Similar to standard DP, it is convenient to work with operators defined on the space of value functions \cite{Ber2012DynamicProgramming}. In our case, Theorem \ref{thm:decompose_CVaR} leads to the following definition of CVaR Bellman operator $\mathbf T : \mathcal X\times\mathcal Y \to \mathcal X\times\mathcal Y$:
\begin{equation}\label{bellman_CVaR}
\mathbf T[V](x,y)= \min_{a\in\mathcal A}\left[C(x,a)+\gamma\max_{\xi\in \U_{\text{CVaR}}(y, P(\cdot|x,a))}\sum_{x'\in\mathcal X}\xi(x')V\left(x',y\xi(x')\right)P(x'|x,a)\right].
\end{equation}
We now establish several useful properties for the Bellman operator $\mathbf T[\V]$.
%These properties will be used in the sequel for proving the correctness of our DP approach.
\vspace{-0.05in}
\begin{lemma}[Properties of CVaR Bellman Operator]\label{lem:prop_Bellman}
The Bellman operator $\mathbf T[\V]$ has the following properties:
\begin{enumerate}
%\item (Monotonicity in $V$) If $\V_1(x,y)\geq \V_2(x,y)$ for any $(x,y)\in\mathcal X \times\mathcal Y$, then $\mathbf T[\V_1](x,y)\geq \mathbf T[\V_2](x,y)$.
%\item (Constant shift) For $K\in\reals$, $\mathbf T[\V+K](x,y)=\mathbf T[\V](x,y)+\gamma K$.
\item (Contraction.) $ \|\mathbf T[\V_1]-\mathbf T[\V_2]\|_{\infty}\leq \gamma \|\V_1-\V_2\|_{\infty}, $ where $\|f\|_{\infty}\!\! = \!\sup_{x\in\X,y\in\mathcal Y} |f(x,y)|$.
\item (Concavity preserving in $y$.) For any $x\in\X$, suppose
$y \V(x,y)$ is concave in $y\in\mathcal Y$. Then the maximization problem in \eqref{bellman_CVaR} is concave. Furthermore, $y \mathbf T[\V](x,y)$ is concave in $y$.
\end{enumerate}
\end{lemma}
\vspace{-0.05in}
The first property in Lemma \ref{lem:prop_Bellman} is similar to standard DP \cite{Ber2012DynamicProgramming}, and is  instrumental to the design of a converging value-iteration approach. The second property is nonstandard and specific to our approach. It will be used to show that the computation of value-iteration updates involves concave, and therefore \emph{tractable} optimization problems. Furthermore, it will be used to show that a linear-interpolation of $V(x,y)$ in the augmented state $y$ has a bounded error.

Equipped with the results in Theorem \ref{thm:decompose_CVaR} and Lemma \ref{lem:prop_Bellman}, we can now  show that the fixed point solution of $\mathbf T[V](x,y)=V(x,y)$ is unique, and equals to the solution of the CVaR MDP problem \eqref{eq:problem} with $x_0=x$ and $\alpha = y$.
\vspace{-0.05in}
\begin{theorem}[Optimality Condition]\label{thm:MDP_CVaR}
For any $x\in\mathcal X$ and $y\in(0,1]$, the solution to $\mathbf T[V](x,y)=V(x,y)$ is unique, and equals to $V^*(x,y) = \min_{\mu\in\Pi_H} \text{CVaR}_y\left(\lim_{T\rightarrow\infty}\mathcal C_{0,T}\mid x_0=x,\mu\right)$.
\end{theorem}
\vspace{-0.05in}
Next, we show that the optimal value of the CVaR MDP problem \eqref{eq:problem} can be attained by a stationary Markov policy, defined as a greedy policy with respect to the value function $V^*(x,y)$. Thus, while the original problem is defined over the intractable space of history-dependent policies, a stationary Markov policy (over the augmented state space) is optimal, and can be readily derived from $V^*(x,y)$. Furthermore, an optimal history-dependent policy can be readily obtained from an (augmented) optimal Markov policy according to the following theorem.
%The original class of feasible policy is history dependent, i.e., $\mu\in\Pi_H$. While searching for a history dependent optimal policy is in general mathematically intractable, by the following theorem we can show that equivalently the optimal policy can be found by optimizing over the class of augmented stationary Markovian policies by robust dynamic programming with respect to the original state $x\in\mathcal X$ and augmented state $y\in\mathcal Y$. This further implies that the class of augmented stationary Markovian policies is \emph{dominating}, i.e., this class of policies is optimal.
\vspace{-0.05in}
\begin{theorem}[Optimal Policies]\label{thm:policy}
Let $\pi_H^*=\{\mu_0,\mu_1,\ldots\}\in\Pi_H$ be a history-dependent policy recursively defined as:
\begin{equation}\label{eq:policy_construct}
\mu_k(h_k) = u^*(x_k, y_k),\,\,\forall k\geq 0,
\end{equation}
with initial conditions $x_0$ and $y_0=\alpha$, and state transitions
\begin{equation}\label{eq:opt_state}
x_k\sim P(\cdot\mid x_{k-1},u^*(x_{k-1},y_{k-1})),\quad y_k = y_{k-1}\xi_{x_{k-1},y_{k-1},u^*}^*(x_k), \forall k\geq 1,
\end{equation}
%\begin{equation}\label{eq:opt_state}
%x_k\sim \xi_{x_{k-1},y_{k-1},u^*}^*(\cdot)P(\cdot\mid x_{k-1},u^*(x_{k-1},y_{k-1})),\,\, y_k = y_{k-1}\xi_{x_{k-1},y_{k-1},u^*}^*(x_k), \forall k\geq 1,
%\end{equation}
where the stationary Markovian policy $u^*(x,y)$ and risk factor $\xi_{x,y,u^*}^*(\cdot)$ are solution to the  min-max optimization problem in the CVaR Bellman operator $\mathbf T[V^*](x,y)$.
Then, $\pi^*_H$ is an optimal policy for problem \eqref{eq:problem} with initial state $x_0$ and CVaR confidence level $\alpha$.
\end{theorem}
\vspace{-0.05in}
%Simply stated, for calculating the optimal policy $u^*$, one needs to \emph{keep track} of the confidence parameter $y$, and update it along the trajectory, according to the observed transitions, using Eq.~\eqref{eq:opt_state}.

Theorems \ref{thm:MDP_CVaR} and \ref{thm:policy} suggest that a value-iteration DP method \cite{Ber2012DynamicProgramming} can be used to solve the CVaR MDP problem \eqref{eq:problem}. Let an initial value-function guess $V_0:\mathcal X\times\mathcal Y\rightarrow\reals$ be chosen arbitrarily. Value iteration proceeds recursively as follows:
\begin{equation}\label{eq:VI}
V_{k+1}(x,y)=\mathbf T[V_k](x,y),\,\forall (x,y)\in\mathcal X\times\mathcal Y,\,k\in\{0,1,\ldots,\}.
\end{equation}
Specifically, by combining the contraction property in Lemma \ref{lem:prop_Bellman} and uniqueness result of fixed point solutions from Theorem \ref{thm:MDP_CVaR}, one concludes that  $\lim_{k\rightarrow\infty}V_k(x,y)=V^*(x,y)$. By selecting $x=x_0$ and $y=\alpha$, one immediately obtains   $V^*(x_0,\alpha)=\min_{\mu\in\Pi_H} \text{CVaR}_{\alpha}\left(\lim_{T\rightarrow\infty}\mathcal C_{0,T}\mid x_0,\mu\right)$. Furthermore, an optimal policy may be derived from $V^*(x,y)$ according to the policy construction procedure  in Theorem \ref{thm:policy}.

Unfortunately, while value iteration is conceptually appealing, its direct implementation in our setting is generally impractical since, e.g., the state $y$ is continuous. In the following, we pursue an \emph{approximation} to the value iteration algorithm \eqref{eq:VI}, based on a linear interpolation scheme for $y$.
\vspace{-0.1in}
\section{Value Iteration with Linear Interpolation}\label{sec:algorithm_interpolation}
\vspace{-0.1in}

In this section we present an approximate DP algorithm for solving CVaR MDPs, based on the theoretical results of Section \ref{subsec:Risk-Bellman}. The value iteration algorithm in Eq. \eqref{eq:VI} presents two main implementation challenges. 
%Theorem \ref{thm:MDP_CVaR} suggests that a value-iteration method, i.e.~equation \eqref{eq:VI}, where one repeatedly applies the Bellman operator $\mathbf T$ to an initial guess of the value function $V_0$ \citep{Ber2012DynamicProgramming}, would converge to the optimal value function. However, compared to standard value iteration \cite{Ber2012DynamicProgramming}, there are two additional difficulties in actually applying $\mathbf T$ in our case. 
The first is due to the fact that the augmented state $y$ is continuous. We handle this challenge by using interpolation, and exploit the concavity of $yV(x,y)$ to bound the error introduced by this procedure. The second challenge stems from the the fact that applying $\mathbf T$ involves maximizing over $\xi$. Our strategy is to exploit the concavity of the maximization problem to guarantee that such optimization can indeed be performed effectively.

As discussed, our approach relies on the fact that the Bellman operator $\mathbf T$ preserves concavity as established in Lemma \ref{lem:prop_Bellman}. Accordingly, we require the following assumption for the initial guess $V_0(x,y)$, 
\begin{assumption}\label{ass:V_0}
The guess for the initial value function $V_0(x,y)$ satisfies the following properties: 1) $y \V_0(x,y)$ is concave in $y\in\mathcal Y$ and 2)  $\V_0(x,y)$ is continuous in $y\in\mathcal Y$ for any $x\in\mathcal X$ .
\end{assumption}
Assumption \ref{ass:V_0} may easily be satisfied, for example, by choosing $V_0(x,y) = \text{CVaR}_{y}(Z\mid x_0=x)$, where $Z$ is any arbitrary bounded random variable.
%\vspace{-0.1in}
\begin{algorithm}[t]
\caption{\texttt{CVaR Value Iteration with Linear Interpolation}}\label{alg:CVI}
1: \alggiven
\begin{itemize}
%\item An interpolation error bound $\epsilon>0$ for small CVaR thresholds.
\item $N(x)$ interpolation points $\interpY(x)  = \left\{y_1,\dots,y_{N(x)}\right\} \in [0,1]^{N(x)}$ for every $x\in \mathcal X$ with $y_i<y_{i+1}$, $y_1=0$ and $y_{N(x)}=1$.
%, $y_2 = \min_{x',a}\{ P(x'|x,a):P(x'|x,a)\neq 0\}$
%\item An interpolation function $\interpI_{x}[V](y;\interpY(x))$ for $yV(x,y)$ for any arbitrary value function $V$.
\item Initial value function $V_0(x,y)$ that satisfies Assumption \ref{ass:V_0}.
 %where $V_0(x,y)=0$ at $y<0$.
\end{itemize}
2: For $t = 1,2,\dots$
\begin{itemize}
%\item Update the smallest non-zero grid $y_2$ in $\interpY(x)$ by choosing it to satisfy $\max_{x\in\mathcal X, y\in \mathbf I_2(x)}|V_0(x,y_2)-V_0(x,y)|\leq\epsilon$, where the interpolation based Bellman operator $ \bellint$ is given by
%  \[
%\hspace{-0.5in}  \bellint[V](x,y) =
% \min_{a\in\mathcal A}\left[C(x,a)+\gamma\max_{\xi\in \U_{\text{CVaR}}(y, P(\cdot|x,a))}\sum_{x'\in\mathcal X}\frac{\interpI_{x'}[V](y\xi(x');\interpY(x'))}{y}P(x'|x,a)\right].
%   \]
\item For each $x \in \mathcal X$ and each $y_i\in \interpY(x)$, update the value function estimate as follows:
  \begin{equation*}
   V_t(x,y_i)= \bellint[V_{t-1}](x,y_i),
  \end{equation*}
  \end{itemize}
3: Set the converged value iteration estimate as $\widehat{V}^*(x,y_i)$, for any $x\in\mathcal X$, and $ y_i\in\interpY(x)$.
\end{algorithm}
%\vspace{-0.1in}
As stated earlier, a key difficulty in applying value iteration \eqref{eq:VI} is that, for each state $x\in \mathcal X$, the Bellman operator has to be calculated for each $y \in \mathcal Y$, and $\mathcal Y$ is continuous. As an approximation, we propose to calculate the Bellman operator only for a finite set of values $y$, and interpolate the value function in between such interpolation points. 

Formally, let $N(x)$ denote the number of interpolation points. For every $x\in\mathcal X$, denote by $\interpY(x) =  \left\{y_1,\dots,y_{N(x)}\right\} \in [0,1]^{N(x)}$ the set of interpolation points. We denote by $\interpI_{x}[V](y)$ the linear interpolation of the function $yV(x,y)$ on these points, i.e.,
\begin{equation*}
\interpI_{x}[V](y)=y_iV(x,y_{i})+\frac{y_{i+1}V(x,y_{i+1})-y_iV(x,y_{i})}{y_{i+1}-y_i}(y-y_i),
\end{equation*}
where $y_i = \max \left\{y'\in \interpY(x) : y' \leq y\right\}$. The interpolation of $yV(x,y)$ instead of $V(x,y)$ is key to our approach. The motivation is twofold: first, it can be shown \cite{rockafellar2000optimization} that for a discrete random variable $Z$, $y \text{CVaR}_{y} (Z)$ is piecewise linear in $y$. Second, one can show that the Lipschitzness of $y\,V(x,y)$ is preserved during value iteration, and exploit this fact to bound the linear interpolation error.
%the definition of $\text{CVaR}_y$ from Theorem 14 of \cite{rockafellar2000optimization} indicates that  This means that $y \text{CVaR}_{y} (Z)$ can be exactly captured by linear interpolation with finite number of segments. However the cut-off points of these segments are unknown in advance. We will later analyze the error bound due to the interpolation and verify that the interpolation error vanishes as the set of interpolation points is sufficiently large.

We now define the \emph{interpolated} Bellman operator $ \bellint$ as follows:
\begin{equation}\label{eq:bell_interp}
\hspace{-0.5in}  \bellint[V](x,y) =
 \min_{a\in\mathcal A}\left[C(x,a)+\gamma\max_{\xi\in \U_{\text{CVaR}}(y, P(\cdot|x,a))}\sum_{x'\in\mathcal X}\frac{\interpI_{x'}[V](y\xi(x'))}{y}P(x'|x,a)\right].
\end{equation}
\vspace{-0.05in}

%Note that the only difference between $ \bellint$ and the original Bellman operator $\mathbf T$, is that $ \bellint$ uses the interpolation of the previous value function.
\vspace{-0.05in}
\begin{remark}
Notice that by L'Hospital's rule one has $\lim_{y\rightarrow 0}{\interpI_{x}[V](y\xi(x))}/{y}=V(x,0)\xi(x)$. This implies that at $y=0$ the interpolated Bellman operator is equivalent to the original Bellman operator, i.e.,
%\begin{equation}\label{TV_0}
$
\mathbf T[V](x,0)= \min_{a\in\mathcal A}\left\{ C(x,a)+\gamma\max_{x'\in\mathcal X : P(x'|x,a)>0}V(x',0) \right\}=\bellint[V](x,0).
$
%\end{equation}
%\begin{equation}\label{TV_0}
%\begin{split}
%\mathbf T[V](x,0)=& \min_{a\in\mathcal A}\left[C(x,a)+\gamma\max_{\xi\in \U_{\text{CVaR}}(0 , P(\cdot|x,a))}\sum_{x'\in\mathcal X}V(x',0)\xi(x')P(x'|x,a)\right]\\
%=& \min_{a\in\mathcal A}\left\{ C(x,a)+\gamma\max_{x'\in\mathcal X : P(x'|x,a)>0}V(x',0) \right\}=\bellint[V](x,0).
%\end{split}
%\end{equation}
\end{remark}
\vspace{-0.05in}

Algorithm \ref{alg:CVI} presents CVaR value iteration with linear interpolation. The only difference between this algorithm and standard value iteration \eqref{eq:VI} is the linear interpolation procedure described above. In the following, we show that Algorithm \ref{alg:CVI} converges, and bound the error due to interpolation.
We begin by showing that the useful properties established in Lemma \ref{lem:prop_Bellman} for the Bellman operator $\mathbf T$ extend to the interpolated Bellman operator $ \bellint$.
\vspace{-0.05in}
\begin{lemma}[Properties of Interpolated Bellman Operator]\label{lem:Bellman_prop_inter}
$\bellint[\V]$ has the same properties of ${\mathbf T}[\V]$ as in Lemma \ref{lem:prop_Bellman}, namely 1) contraction and 2) concavity preservation.
\end{lemma}
\vspace{-0.05in}

Lemma \ref{lem:Bellman_prop_inter} implies several important consequences for Algorithm \ref{alg:CVI}. The first one is that the maximization problem in \eqref{eq:bell_interp} is concave, and thus may be solved efficiently at each step. This guarantees that the algorithm is \emph{tractable}.
Second, the contraction property in Lemma \ref{lem:Bellman_prop_inter} guarantees that Algorithm \ref{alg:CVI} converges, i.e., there exists a value function $\widehat{V}^*\in\reals^{|\mathcal X|\times|\mathcal Y|}$ such that $\lim_{n\rightarrow\infty}\bellint^n[V_{0}](x,y_i)=\widehat{V}^*(x,y_i)$. In addition, the convergence rate is geometric and equals to $\gamma$.

%We now investigate the error due to the linear interpolation in Algorithm \ref{alg:CVI}.
%We shall  relate the error bound $\delta$ with the number of interpolation points $N(x)$. In general, on top of showing a consistency result, i.e., $\delta\rightarrow 0$ when $N(x)\rightarrow\infty$ for any $x\in\mathcal X$, we also want to find the convergence rate of the error bound with respect to $N(x)$.
The following theorem provides an error bound between approximate value iteration and exact value iteration  \eqref{eq:problem} in terms of the interpolation resolution.
\vspace{-0.05in}
 \begin{theorem}[Convergence and Error Bound]\label{thm:interpolation_bdd}
Suppose the initial value function $V_0(x,y)$ satisfies Assumption \ref{ass:V_0} and let $\epsilon>0$ be an error tolerance parameter. For any state $x\in\mathcal X$ and step $t\geq 0$, choose $y_2>0$ such that $V_t(x,y_2)- V_t(x,0)\geq -\epsilon$ and update the interpolation points according to the logarithmic rule: $y_{i+1}=\theta y_i$, $\forall i\geq 2$, with uniform constant $\theta\geq 1$. Then, Algorithm \ref{alg:CVI} has the following error bound:
\begin{small} \[
0\geq \widehat{V}^*(x_0,\alpha)-\min_{\mu\in\Pi_H} \text{CVaR}_{\alpha}\left(\lim_{T\rightarrow\infty}\mathcal C_{0,T}\mid x_0,\mu\right)\geq\frac{-\gamma}{1-\gamma}\mathbf{O}\left((\theta-1)+\epsilon\right),
\]\end{small}
 and the following finite time convergence error bound:
\begin{small} \[
 \left|\bellint^n[V_0](x_0,\alpha)-\min_{\mu\in\Pi_H} \text{CVaR}_{\alpha}\left(\lim_{T\rightarrow\infty}\mathcal C_{0,T}\mid x_0,\mu\right)\right|\leq\frac{\mathbf{O}\left((\theta-1)+\epsilon\right)+\mathbf{O}(\gamma^n)}{1-\gamma}.
\]\end{small}
 \end{theorem}
 \vspace{-0.05in}
Theorem \ref{thm:interpolation_bdd} shows that 1) the interpolation-based value function is a \emph{conservative estimate} for the optimal solution to problem \eqref{eq:problem}; 2) the interpolation procedure is \emph{consistent}, i.e., when the number of interpolation points is arbitrarily large (specifically, $\epsilon\rightarrow 0$ and ${y_{i+1}}/{y_{i}}\rightarrow 1$), the approximation error tends to zero; and 3) the approximation error bound is $O((\theta-1) + \epsilon)$, where $\log\theta$ is the \emph{log-difference} of the interpolation points, i.e., $\log\theta= \log y_{i+1}-\log y_i$, $\forall i$.

For a pre-specified $\epsilon$, the condition $V_t(x,y_2)- V_t(x,0)\geq -\epsilon$ may be satisfied by a simple \emph{adaptive procedure} for selecting the interpolation points $\interpY(x)$. At each iteration $t > 0$, after calculating $V_t(x,y_i)$ in Algorithm \ref{alg:CVI}, at each state $x$ in which the condition does not hold, add a new interpolation point $y_2' = \frac{\epsilon y_2}{|V_t(x,y_2) - V_t(x,0)|}$, and additional points between $y_2'$ and $y_2$ such that the condition $\log\theta \geq \log y_{i+1}-\log y_i$ is maintained. Since all the additional points belong to the segment $[y_1,y_2]$, the linearly interpolated $V_t(x,y_i)$ remains unchanged, and Algorithm \ref{alg:CVI} proceeds as is. For bounded costs and $\epsilon > 0$, the number of additional points required is bounded.

The full proof of Theorem \ref{thm:interpolation_bdd} is detailed in the supplementary material; we highlight the main ideas and challenges involved.
In the first part of the proof we bound, for all $t>0$, the Lipschitz constant of $y V_t(x,y)$ in $y$. The key to this result is to show that the Bellman operator $\mathbf T$ preserves the Lipschitz property for $y V_t(x,y)$. Using the Lipschitz bound and the concavity of $y V_t(x,y)$, we then bound the error $\frac{\interpI_{x}[V_{t}](y)}{y}-V_{t}(x,y)$ for all $y$. The condition on $y_2$ is required for this bound to hold when $y \to 0$. Finally, we use this result to bound $\| \bellint[V_t](x,y)- {\mathbf T}[V_t](x,y) \|_\infty$. The results of Theorem \ref{thm:interpolation_bdd} follow from contraction arguments, similar to approximate dynamic programming \cite{Ber2012DynamicProgramming}.

\vspace{-0.15in}
\section{Experiments}\label{sec:experiments}
\vspace{-0.1in}
We validate Algorithm \ref{alg:CVI} on a rectangular grid world,
where states represent grid points on a 2D terrain map. An agent (e.g., a robotic vehicle) starts in a safe region and its objective is to travel  to a given  destination. 
At each time step the agent can move to any of its four neighboring states. Due to sensing and control noise, however, with probability $\delta$ a move to a random neighboring state occurs. The stage-wise cost of each move until reaching the destination is $1$, to account for  fuel usage.  In between the starting point and the destination there are a number of obstacles that the agent should avoid.  Hitting an obstacle costs $M>>1$ and terminates the mission. The objective is to compute a \emph{safe} (i.e., obstacle-free) path that is \emph{fuel efficient}. 

%We used Algorithm \ref{alg:CVI} to find a risk-sensitive policy, that trades-off safety and fuel efficiency.

%In order to trade-off between safety and fuel efficiency for the whole mission, here we choose to minimize the $\text{CVaR}$ risk of the total (discounted) cost function, where the discount factor $\gamma$ is designed to capture the effective planning horizon $1/(1-\gamma)$. We then apply Algorithm \ref{alg:CVI} to solve for risk sensitive optimal policies and evaluate their robustness to model perturbations.

\begin{figure}
\begin{center}
  \includegraphics[width=\textwidth]{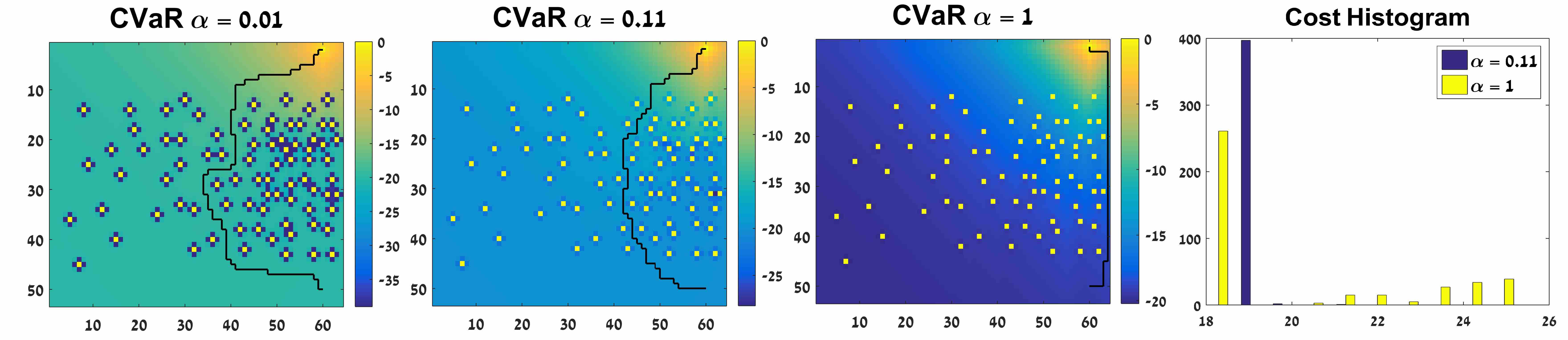}
  \vspace{-0.15in}
  \caption{Grid-world simulation. Left three plots show the value functions and corresponding paths  for different CVaR confidence levels. The rightmost plot shows a cost histogram (for 400 Monte Carlo trials) for a risk-neutral policy and a CVaR policy with confidence level $\alpha = 0.11$.}\label{fig:gridworld1}
\end{center}
\vspace{-0.2in}
\end{figure}

For our experiments, we choose a $64 \times 53$ grid-world (see Figure \ref{fig:gridworld1}),
%in Figure \ref{fig:gridworld1} where the trade-off between safety and efficient path follows from the solution of the CVaR MDP with differ confidence intervals. Here our
for a total of 3,312 states. The destination is at position $(60,2)$, and there are $80$ obstacles plotted in yellow. By leveraging Theorem \ref{thm:interpolation_bdd}, we  use $21$ log-spaced interpolation points for Algorithm \ref{alg:CVI} in order to achieve a small value function error.
%The random noise in choosing a wrong action, i.e. an action that moves to a random direction instead of the intended direction, is with probability
We choose $\delta=0.05$, and a discount factor $\gamma = 0.95$ for an effective horizon of 200 steps. Furthermore, we set the penalty cost equal to $M=2/(1-\gamma)$--such choice trades off high  penalty for collisions and computational complexity (that increases as $M$ increases).

%At all states that are not adjacent to an obstacle, the transitions are deterministic, and correspond to the chosen action. At states adjacent to an obstacle, a collision occurs with probability $\epsilon$, following which the agent transitions to the initial position at $(12,13)$. If a collision does not occur, the agent transitions according to his chosen action.
%The cost at each state is $1$, regardless of the actions or collisions, and the target state is absorbing, with zero reward.

In Figure \ref{fig:gridworld1} we plot the value function $V(x,y)$ for three different values of the CVaR confidence parameter $\alpha$, and the corresponding paths  starting from the initial position $(60,50)$. The first three figures in Figure \ref{fig:gridworld1} show how by decreasing the confidence parameter $\alpha$ the average travel distance (and hence fuel consumption) slightly increases but the collision probability decreases, as expected.  We next discuss robustness to modeling errors. We conducted simulations in which with probability $0.5$ each obstacle position is perturbed in a random direction to one of the neighboring grid cells. This emulates, for example, measurement errors in the terrain map. We then trained both the risk-averse ($\alpha=0.11$) and risk-neutral ($\alpha = 1$) policies on the nominal (i.e., unperturbed) terrain map, and evaluated them on $400$ perturbed scenarios ($20$ perturbed maps with $20$ Monte Carlo evaluations each).  While the risk-neutral policy finds a shorter route (with average cost equal to $18.137$ on successful runs), it is vulnerable to perturbations and fails more often (with over $120$ failed runs). In contrast, the risk-averse policy chooses slightly longer routes (with average cost equal to $18.878$ on successful runs), but is much more robust to model perturbations (with only $5$ failed runs).

For the computation of Algorithm \ref{alg:CVI} we represented the concave piecewise linear maximization problem in \eqref{eq:bell_interp} as a linear program, and concatenated several problems to reduce repeated overhead stemming from the initialization of the CPLEX linear programming solver. This resulted in a computation time on the order of two hours. We believe there is ample room for improvement, for example by leveraging parallelization and sampling-based methods. Overall, we believe our proposed approach is currently the most practical method available for solving CVaR MDPs (as a comparison, the recently proposed method in \cite{haskell2014convex} involves infinite dimensional optimization).  The Matlab code used for the experiments  is provided in the supplementary material. 
\vspace{-0.1in}
\section{Conclusion}\label{sec:conclusion}
\vspace{-0.1in}
In this paper we presented an algorithm for CVaR MDPs, based on approximate value-iteration on an augmented state space. We established convergence of our algorithm, and derived finite-time error bounds. These bounds are useful to stop  the algorithm at a desired error threshold.

In addition, we uncovered an interesting relationship between the CVaR of the total cost and the worst-case expected cost under adversarial model perturbations. In this formulation, the perturbations are correlated in time, and lead to a robustness framework significantly less conservative than the popular robust-MDP framework, where the uncertainty is temporally independent. 

Collectively, our work suggests CVaR MDPs as a unifying and practical framework for computing control policies that are robust with respect to both stochasticity and model perturbations. Future work should address extensions  to large state-spaces. We conjecture that a sampling-based approximate DP approach \cite{Ber2012DynamicProgramming} should be feasible since, as proven in this paper, the CVaR Bellman equation is contracting (as required by approximate DP methods). 
%\vspace{-0.15in}
%\section{Acknowledgement}\label{sec:acknowledgement}
%\vspace{-0.1in}
%The authors would like to thank Mohammad Ghavamzadeh for his comments that helped us with some technical details of the interpolation based DP algorithm.
% comment this part if needed for anonymity in review

\newpage
\begin{small}
\bibliography{SpectralRiskNIPS15}
\bibliographystyle{plainnat}
\end{small}

\newpage
\appendix
\section{Proofs of Theoretical Results}
\subsection{Proof of Proposition \ref{prop:CVaR_is_robustness}}
By definition, we have that
\begin{equation*}
\begin{split}
    \ExpW{C(x_1,\dots,x_T)}{\hat{P}} &= \sum_{(x_1,\dots,x_T)} P_1(x_1)\delta_1(x_1)\cdots P_T(x_T|x_{T-1})\delta_T(x_T|x_{T-1})C(x_1,\dots,x_T) \\
    &= \sum_{(x_1,\dots,x_T)} \pr(x_1,\dots,x_T) \delta_1(x_1)\delta_2(x_2|x_1)\cdots \delta_T(x_T|x_{T-1}) C(x_1,\dots,x_T)\\
    &\doteq \sum_{(x_1,\dots,x_T)} \pr(x_1,\dots,x_T) \delta(x_1,\dots,x_T) C(x_1,\dots,x_T).
\end{split}
\end{equation*}
Note that by definition of the set $\Delta$, for any $(\delta_1,\dots,\delta_T)\in \Delta$ we have that $\pr(x_1,\dots,x_T)>0 \rightarrow \delta(x_1,\dots,x_T) \geq 0$, and
\begin{equation*}
    \Exp{\delta(x_1,\dots,x_T)} \doteq \sum_{(x_1,\dots,x_T)} \pr(x_1,\dots,x_T) \delta(x_1,\dots,x_T) = 1.
\end{equation*}
Thus,
\begin{equation*}
\begin{split}
    \sup_{(\delta_1,\dots,\delta_T)\in \Delta_{\eta}} \ExpW{C(x_1,\dots,x_T)}{\hat{P}} &= \sup_{\substack{0\leq\delta(x_1,\dots,x_T)\leq \eta,\\ \Exp{\delta(x_1,\dots,x_T)}=1}} \sum_{(x_1,\dots,x_T)} \pr(x_1,\dots,x_T) \delta(x_1,\dots,x_T) C(x_1,\dots,x_T) \\
    &= \text{CVaR}_{\frac{1}{\eta}} \left(C(x_1,\dots,x_T)\right),
\end{split}
\end{equation*}
where the last equality is by the representation theorem for CVaR \cite{Shapiro2009}.

\subsection{Proof of Lemma \ref{lem:prop_Bellman}}
The proof of monotonicity and constant shift properties follow directly from the definitions of the Bellman operator, by noting that $\xi(x')P(x'|x,a)$ is non-negative and $\sum_{x'\in\mathcal X}\xi(x')P(x'|x,a)]=1$ for any $\xi\in \U_{\text{CVaR}}(y, P(\cdot|x,a))$.
For the contraction property, denote $c=\|\V_1-\V_2\|_{\infty}$. Since
\[
\V_2(x,y)-\|\V_1-\V_2\|_{\infty}\leq \V_1(x,y)\leq \V_2(x,y)+\|\V_1-\V_2\|_{\infty},\,\, \forall x\in\X, \,y\in\mathcal Y,
\]
by monotonicity and constant shift property,
\[
\mathbf T[\V_2](x,y)-\gamma\|\V_1-\V_2\|_{\infty}\leq \mathbf T[\V_1](x,y) \leq \mathbf T[\V_2](x,y)+\gamma\|\V_1-\V_2\|_{\infty}\,\, \forall x\in\X, \,y\in\mathcal Y.
\]
This further implies that
\[
|\mathbf T[\V_1](x,y)-\mathbf T[\V_2](x,y)|\leq \gamma\|\V_1-\V_2\|_{\infty}\,\, \forall x\in\X, \,y\in\mathcal Y
\]
and the contraction property follows.

Now, we prove the concavity preserving property. Assume that $y \V(x,y)$ is concave in $y$ for any $x\in\X$. Let $y_1,y_2 \in\mathcal Y$, and $\lambda\in[0,1]$, and define $y_\lambda = (1-\lambda) y_1 + \lambda y_2$. We have
\begin{equation*}
\begin{split}
  & (1-\lambda)y_1 \mathbf T [\V](x,y_1) + \lambda y_2 \mathbf T [\V](x,y_2) \\
  = & (1-\lambda)y_1 \min_{a_1\in\mathcal A}\left[C(x,a_1)+\gamma\max_{\xi_1\in \U_{\text{CVaR}}(y_1, P(\cdot|x,a_1))}\sum_{x'\in\mathcal X}\xi_1(x')V\left(x',y_1 \xi_1(x')\right)P(x'|x,a_1)\right]\\
  & + \lambda y_2 \min_{a_2\in\mathcal A}\left[C(x,a_2)+\gamma\max_{\xi_2\in \U_{\text{CVaR}}(y_2, P(\cdot|x,a_2))}\sum_{x'\in\mathcal X}\xi_2(x')V\left(x',y_2 \xi_2(x')\right)P(x'|x,a_2)\right]\\
  = & \min_{a_1\in\mathcal A}\left[(1-\lambda)y_1 C(x,a_1)+\gamma\max_{\xi_1\in \U_{\text{CVaR}}(y_1, P(\cdot|x,a_1))}\sum_{x'\in\mathcal X}\xi_1(x')V\left(x',y_1 \xi_1(x')\right)P(x'|x,a_1)(1-\lambda)y_1 \right]\\
  & + \min_{a_2\in\mathcal A}\left[\lambda y_2 C(x,a_2)+\gamma\max_{\xi_2\in \U_{\text{CVaR}}(y_2, P(\cdot|x,a_2))}\sum_{x'\in\mathcal X}\xi_2(x')V\left(x',y_2 \xi_2(x')\right)P(x'|x,a_2)\lambda y_2 \right]
  \end{split}
  \end{equation*}
  \begin{equation*}
\begin{split}
  \leq & \min_{a\in\mathcal A}\left[y_\lambda C(x,a)+\gamma \!\!\!\!\!\! \max_{\substack{\xi_1\in \U_{\text{CVaR}}(y_1, P(\cdot|x,a))\\\xi_2\in \U_{\text{CVaR}}(y_2, P(\cdot|x,a))}}\sum_{x'\in\mathcal X}P(x'|x,a)\left((1 \!-\! \lambda)y_1 \xi_1(x')V\left(x',y_1 \xi_1(x')\right) + \lambda y_2 \xi_2(x')V\left(x',y_2 \xi_2(x')\right)\right) \right]\\
  \leq & \min_{a\in\mathcal A}\left[y_\lambda C(x,a)+\gamma \!\!\!\!\!\!\!\!\!\! \max_{\substack{\xi_1\in \U_{\text{CVaR}}(y_1, P(\cdot|x,a))\\\xi_2\in \U_{\text{CVaR}}(y_2, P(\cdot|x,a))}} \! \sum_{x'\in\mathcal X} \!\! P(x'|x,a)\left((1 \!-\! \lambda)y_1 \xi_1(x') \!+\! \lambda y_2 \xi_2(x') \right)V\left(x',((1\!-\!\lambda) y_1 \xi_1(x') + \lambda y_2 \xi_2(x'))\right) \right]\\
\end{split}
\end{equation*}
where the first inequality is by concavity of the $\min$, and the second is by the concavity assumption. Now, define $\xi = \frac{(1\!-\!\lambda) y_1 \xi_1 + \lambda y_2 \xi_2}{y_\lambda}$. When $\xi_1\in \U_{\text{CVaR}}(y_1, P(\cdot|x,a))$ and $\xi_2\in \U_{\text{CVaR}}(y_2, P(\cdot|x,a))$, we have that $\xi \in \left[ 0, \frac{1}{y_\lambda}\right]$ and $\sum_{x'\in\mathcal X} \xi(x')\mathbb P(x'|x,a) =1$. We thus have
\begin{equation*}
\begin{split}
  & (1-\lambda)y_1 \mathbf T [\V](x,y_1) + \lambda y_2 \mathbf T [\V](x,y_2) \\
  \leq & \min_{a\in\mathcal A}\left[y_\lambda C(x,a)+\gamma \max_{\xi \in \U_{\text{CVaR}}(y_\lambda, P(\cdot|x,a))} \sum_{x'\in\mathcal X} P(x'|x,a) y_\lambda \xi(x') V\left(x',y_\lambda \xi(x') \right) \right]\\
  = & y_\lambda \min_{a\in\mathcal A}\left[C(x,a)+\gamma \max_{\xi \in \U_{\text{CVaR}}(y_\lambda, P(\cdot|x,a))} \sum_{x'\in\mathcal X} P(x'|x,a) \xi(x') V\left(x',y_\lambda \xi(x') \right) \right] =  y_\lambda \mathbf T[\V](x,y_\lambda).
\end{split}
\end{equation*}

Finally, to show that the inner problem in \eqref{bellman_CVaR} is a concave maximization, we need to show that
\[
\Lambda_{x,y,a}(z):=\left\{\begin{array}{cl}
zV(x',z){P(x'|x,a)}/{y}&\text{if $y\neq 0$}\\
0&\text{otherwise}
\end{array}\right.
\]
is a concave function in $z\in\reals$ for any given $x\in\mathcal X$, $y\in\mathcal Y$ and $a\in\mathcal A$.
Suppose $zV(x,z)$ is a concave function in $z$. Immediately we can see that $\Lambda_{x,y,a}(z)$ is concave in $z$ when $y=0$. Also notice that when $y\in\mathcal Y\setminus\{0\}$, since the transition probability $P(x'|x,a)$ is non-negative, we have the result that $\Lambda_{x,y,a}(z)$ is concave in $z$. This further implies
\[
\sum_{x'\in\mathcal X}\frac{P(x'|x,a)}{y}\Lambda_{x,y,a}(y\xi(x'))=\sum_{x'\in\mathcal X}\xi(x')V(x',y\xi(x'))P(x'|x,a)
\]
is concave in $\xi$. Furthermore by combining the result with the fact that the feasible set of $\xi$ is a polytope, we complete the proof of this claim.

\subsection{Proof of Theorem \ref{thm:MDP_CVaR}}
The first part of the proof is to show that for any $(x,y)\in\X\times\mathcal Y$,
\begin{equation}\label{eq:ind}
V_n(x,y):=\mathbf{T}^{n}[V_{0}](x,y)\!=\!\min_{\mu\in\Pi_{M}}\text{CVaR}_y\left(\mathcal C_{0,n}+\gamma^nV_0\mid x_0=x,\mu\right),
\end{equation}
by induction, where the initial condition is $(x_0,y_0)=(x,y)$ and control action $a_t$ is induced by $\mu(x_t,y_t)$. For $n=1$, we have that $V_1(x,y)=\mathbf{T}[V_{0}](x,y)=\min_{\mu\in\Pi_{M}}{C}(x_0,a_0)+\gamma \text{CVaR}_y\left(C(x_1,a_1)+V_0(x_1)\mid x_0=x,\mu\right)$ from definition.
By induction hypothesis, assume the above expression holds at $n=k$. For $n=k+1$,
\begin{equation}\label{eq:part1}
\begin{split}
&V_{k+1}(x,y):=\mathbf{T}^{k+1}[V_{0}](x,y)=\mathbf{T}[V_k](x,y)\\
=&\min_{a\in\mathcal A}\left[C(x,a)+\gamma\max_{\xi\in \U_{\text{CVaR}}(y, P(\cdot|x,a))}\sum_{x'\in\mathcal X}\xi(x')V_k\bigg(x',\underbrace{y\xi(x')}_{y'}\bigg)P(x'|x,a)\right]\\
=& \min_{a\in\mathcal A}\Bigg[{C}(x,a)+\gamma\max_{\xi\in \U_{\text{CVaR}}(y, P(\cdot|x,a))}\sum_{x'\in\mathcal X}\xi(x')P(x'|x,a)\min_{\mu\in\Pi_{M}}
\text{CVaR}_{y'}\left(\mathcal C_{0,k}+\gamma^kV_0\mid x_0=x',\mu\right)\Bigg]\\
=&\min_{a\in\mathcal A}\left[{C}(x,a)+\max_{\xi\in \U_{\text{CVaR}}(y, P(\cdot|x,a))}\mathbb E_{\xi}\bigg[\min_{\mu\in\Pi_{M}}
\text{CVaR}_{y_1}\left(\mathcal C_{1,k+1}+\gamma^{k+1}V_0\mid x_1,\mu\right)\bigg]\right]\\
=&\min_{\mu\in\Pi_M}
\text{CVaR}_y\left(\mathcal C_{0,k+1}+\gamma^{k+1}V_0\mid x_0=x,\mu\right),
\end{split}
\end{equation}
where the initial state condition is given by $(x_0,y_0)=(x,y)$.
Thus, the equality in \eqref{eq:ind} is proved by induction.

The second part of the proof is to show that $V^*(x_0,y_0)\!=\!\min_{\mu\in\Pi_M}
\text{CVaR}_{y_0}\left(\lim_{n\rightarrow\infty} \mathcal C_{0,n}\mid x_0,\mu\right)$.
Recall $\mathbf T[V](x,y) =\min_{a\in\mathcal A}C(x,a) + \gamma\!\max_{\xi\in\U_{\text{CVaR}}(y,P(\cdot|x,a))}\mathbb E_{\xi}[V\!\mid\! x,y,a]$. Since $\mathbf T$ is a contraction and $V_0$ is bounded, one obtains
\[
V^*(x,y)=\mathbf{T}[V^*](x,y)=\lim_{n\rightarrow\infty} \mathbf{T}^{n}[V_0](x,y)=\lim_{n\rightarrow\infty} V_{n}(x,y)
\]
for any $(x,y)\in\mathcal X\times\mathcal Y$.
The first and the second equality follow directly from Proposition 2.1 and Proposition 2.2 in \cite{Ber2012DynamicProgramming} and the third equality follows from the definition of $V_n$.  Furthermore since $V_0(x,y)$ is bounded for any $(x,y)\in\X\times\mathcal Y$,  the result in \eqref{eq:part1} implies
\[
-\lim_{n\rightarrow\infty}\gamma^{n}\|V_0\|_\infty \leq  V^*(x_0,y_0)-\min_{\mu\in\Pi_M}
\text{CVaR}_{y_0}\left(\lim_{n\rightarrow\infty}\mathcal C_{0,n}\mid x_0,\mu\right)\leq \lim_{n\rightarrow\infty}\gamma^{n}\|V_0\|_\infty.
\]
Therefore, by taking $n\rightarrow \infty$, we have just shown that for any $(x_0,y_0)\in\X\times\mathcal Y$, $V^*(x_0,y_0)= \min_{\mu\in\Pi_M}
\text{CVaR}_{y_0}\left(\lim_{n\rightarrow\infty}\mathcal C_{0,n}\mid x_0,\mu\right)$.

The third part of the proof is to show that for the initial state $x_0$ and confidence interval $y_0$, we have that
\[
V^*(x_0,y_0)=\min_{\mu\in\Pi_H}
\text{CVaR}_{y_0}\left(\lim_{n\rightarrow\infty} \mathcal C_{0,n}\mid x_0,\mu\right).
\]
At any $(x_t,y_t)\in\mathcal X\times\mathcal Y$, we first define the $t^{\text{th}}$ tail-subproblem of problem \eqref{eq:problem} as follows:
\[
 \mathbb{V}(x_t,y_t)\!=\! \min_{\mu\in\Pi_H}
\text{CVaR}_{y_t}\left(\lim_{n\rightarrow\infty} \mathcal C_{t,n}\mid x_t,\mu\right)
\]
where the tail policy sequence is equal to $\mu=\{\mu_t,\mu_{t+1},\ldots\}$ and the action is given by $a_j=\mu_j(h_j)$ for $j\geq t$. For any history depend policy $\widetilde\mu\in\Pi_H$, we also define the $\widetilde\mu-$induced value function as $\text{CVaR}_{y_t}\left(\lim_{n\rightarrow\infty} \mathcal C_{t,n}\mid x_t,\widetilde\mu\right)$ where $\widetilde\mu=\{\widetilde\mu_t,\widetilde\mu_{t+1},\ldots\}$ and $a_j=\widetilde\mu_j(h_j)$ for $j\geq t$.

 Now let $\mu^*$ be the optimal policy of the above $t^{\text{th}}$ tail-subproblem.
Clearly, the truncated policy $\widetilde\mu=\{\mu^*_{t+1},\mu^*_{t+2},\ldots\}$ is a feasible policy for the $(t+1)^{\text{th}}$ tail subproblem at any state $x_{t+1}$ and confidence interval $y_{t+1}$:
\[
\min_{\mu\in\Pi_H}
\text{CVaR}_{y_{t+1}}\left(\lim_{n\rightarrow\infty} \mathcal C_{t+1,n}\mid x_{t+1},\mu\right).
\]
Collecting the above results, for any pair $(x_t,y_t)\in\mathcal X\times\mathcal Y$ and with $a_t= \mu_t^*(x_t)$ we can write
\begin{equation*}
\begin{split}
 \mathbb{V}(x_t, y_t) =&C(x_t,a_t)+\gamma\max_{\xi\in \U_{\text{CVaR}}(y_t, P(\cdot|x_t,a_t))}\mathbb E\left[\xi(x_{t+1})\cdot\underbrace{\text{CVaR}_{y_{t+1}}\left(\lim_{n\rightarrow\infty} \mathcal C_{t+1,n}\mid x_{t+1},\widetilde\mu\right)}_{ \mathbb{V}^{ \widetilde\mu}(x_{t+1},y_{t+1}),y_{t+1}=y_t\xi(x_{t+1})} \right]\\
\geq &C(x_t, a_t) \!+\!\gamma\max_{\xi\in \U_{\text{CVaR}}(y_t, P(\cdot|x_t,a_t))}\mathbb E_{\xi}[ \mathbb{V}(x_{t+1},y_t\xi(x_{t+1}))\!\mid\! x_t,y_t,a_t]\!\geq\!  \mathbf T[ \mathbb{V}](x_t, y_t).
\end{split}
\end{equation*}
The first equality follows from the definition of $ \mathbb{V}(x_t,y_t)$ and the decomposition of CVaRs (Theorem \ref{thm:decompose_CVaR}). The first inequality uses the inequality: $ \mathbb{V}^{ \widetilde\mu}(x,y)\geq  \mathbb{V}(x,y)$, $\forall (x,y)\in\mathcal X\times\mathcal Y$. The second inequality follows from the definition of  Bellman operator $\mathbf T$.

On the other hand, starting at any state $x_{t+1}$ and confidence interval $y_{t+1}$, let $\mu^*=\{\mu^*_{t+1},\mu^*_{t+2},\ldots\} \in \Pi_{H}$ be an optimal policy for the tail subproblem:
\[
\min_{\mu\in\Pi_H}\text{CVaR}_{y_{t+1}}\left(\lim_{n\rightarrow\infty} \mathcal C_{t+1,n}\mid x_{t+1},\mu\right).
\]
For a given pair of $(x_t,y_t)\in\mathcal X\times\mathcal Y$, construct  the ``extended" policy $\widetilde\mu=\{\widetilde\mu_t,\widetilde\mu_{t+1},\ldots\} \in \Pi_H$ as follows:
\[
\widetilde \mu_t(x_t) = u^*(x_t,y_t), \text{ and } \widetilde\mu_j(h_{j}) = \mu^*_j(h_{j}) \text{ for } j\geq t+1,
\]
where $u^*(x_t,y_t)$ is the minimizer of the fixed point equation
\[
u^*(x_t,y_t)\in\argmin_{a\in\mathcal A}C(x_t,a) + \gamma\!\max_{\xi\in \U_{\text{CVaR}}(y_t, P(\cdot|x_t,a))}\mathbb E_{\xi}[\mathbb{V}(x_{t+1},y_t\xi(x_{t+1}))\!\mid\! x_t,y_t,a],
\]
with $y_t$ is the given confidence interval to the $t^{\text{th}}$ tail-subproblem and the transition from $y_t$ to $y_{t+1}$ is given by $y_{t+1}=y_t\xi^*(x_{t+1})$ where
\[
\xi^*\in\arg\max_{\xi\in \U_{\text{CVaR}}(y_t, P(\cdot|x_t,a^*))}\mathbb E\left[\xi(x_{t+1})\cdot{\text{CVaR}_{y_t\xi(x_{t+1})}\left(\lim_{n\rightarrow\infty} \mathcal C_{t+1,n}\mid x_{t+1},\widetilde\mu\right)} \right]
\]
Since $\mu^*$ is an optimal, and a fortiori feasible policy for the tail subproblem (from time $t+1$), the policy $\widetilde\mu \in \Pi_H$ is a feasible policy for the tail subproblem (from time $t$): $\min_{\mu\in\Pi_H}\text{CVaR}_{y_{t}}\left(\lim_{n\rightarrow\infty} \mathcal C_{t,n}\mid x_{t},\mu\right)$.
Hence, we can write
\begin{equation*}
\begin{split}
&\mathbb{V}(x_t, y_t) \leq  C(x_t, \widetilde \mu_t(x_t)) + \gamma\text{CVaR}_{y_{t}}\left(\lim_{n\rightarrow\infty} \mathcal C_{t+1,n}\mid x_{t},\widetilde\mu\right).
\end{split}
\end{equation*}
Hence from the definition of $\mu^*$, one easily obtains:
\begin{equation*}
\begin{split}
&\mathbb{V}(x_t, y_t) \\
\leq& C(x_t, u^*(x_t,y_t)) +  \gamma\!\max_{\xi\in \U_{\text{CVaR}}(y_t, P(\cdot|x_t,u^*(x_t,y_t)))}\mathbb E\left[\xi(x_{t+1})\cdot\text{CVaR}_{y_t\xi(x_{t+1})}\left(\lim_{n\rightarrow\infty} \mathcal C_{t+1,n}\mid x_{t+1},\widetilde\mu\right) \mid\! x_t,y_t,u^*(x_t,y_t)\right]
\\
=&  C(x_t, u^*(x_t,y_t)) +  \gamma\!\max_{\xi\in \U_{\text{CVaR}}(y_t, P(\cdot|x_t,u^*(x_t,y_t)))}\mathbb E_{\xi}[\mathbb{V}(x_{t+1},y_t\xi(x_{t+1}))\!\mid\! x_t,y_t,u^*(x_t,y_t)]\\
=& \mathbf T[\mathbb{V}](x_t, y_t).
\end{split}
\end{equation*}

Collecting the above results, we have shown that $\mathbb{V}$ is a fixed point solution to $V(x,y)=\mathbf T[V](x,y)$ for any $(x,y)\in\mathcal X\times\mathcal Y$. Since the fixed point solution is unique, combining both of these arguments implies $V^*(x,y)=\mathbb{V}(x,y)$ for any $(x,y)\in\mathcal X\times\mathcal Y$. Therefore, it follows that with initial state $(x,y)$, we have $V^*(x,y)=\mathbb{V}(x, y) =\min_{\mu\in\Pi_H} \text{CVaR}_y\left(\lim_{T\rightarrow\infty}\mathcal C_{0,T}\mid x_0=x,\mu\right)$.

Combining the above three parts of the proof, the claims of this theorem follows.

\subsection{Proof of Theorem \ref{thm:policy}}
Similar to the definition of the optimal Bellman operator $\mathbf T$, for any augmented stationary Markovin policy $u:\mathcal X\times\mathcal Y\rightarrow\mathcal A$, we define the policy induced Bellman operator $\mathbf T_{u}$ as
\[
\mathbf T_u[V](x,y)= C(x,u(x,y))+\gamma\max_{\xi\in \U_{\text{CVaR}}(y, P(\cdot|x,u(x,y)))}\sum_{x'\in\mathcal X}\xi(x')V\left(x',y\xi(x')\right)P(x'|x,u(x,y)).
\]
Analogous to Theorem \ref{thm:MDP_CVaR}, we can easily show that the fixed point solution to $\mathbf T_u[V](x,y)=V(x,y)$ is unique and the CVaR decomposition theorem (Theorem \ref{thm:decompose_CVaR}) further implies this solution equals to
\[
\text{CVaR}_y\left(\lim_{T\rightarrow\infty}\mathcal C_{0,T}\mid x_0=x,u_H\right),
\]
where the history dependent policy $\pi_H=\{\mu_0,\mu_1,\ldots\}$ is given by $\mu_k(h_k) = u(x_k, y_k)$ for any $k\geq 0$,
with initial states $x_0,y_0=\alpha$, state transitions \eqref{eq:opt_state}, but with augmented stationary Markovian policy $u^*$ replaced by $u$.

To complete the proof of this theorem, we need to show that the augmented stationary Markovian policy $u^*$ is optimal if and only if
\begin{equation}\label{eq:opt_eq}
\mathbf T[V^*](x,y)=\mathbf T_{u^*}[V^*](x,y),\,\,\forall x\in\mathcal X,\,\,y\in\mathcal Y,
\end{equation}
where $V^*(x,y)$ is the unique fixed point solution of $\mathbf T[V](x,y)=V(x,y)$. Here an augmented stationary Markovian policy $u^*$ is optimal if and only if the  induced history dependent policy $u^*_H$ in \eqref{eq:policy_construct} is optimal to problem \eqref{eq:problem}.

First suppose $u^*$ is an optimal augmented stationary Markvoian policy.
Then using the definition of $u^*$ and the result from Theorem \ref{thm:MDP_CVaR} that
\[
V^*(x,y) = \min_{\mu\in\Pi_H} \text{CVaR}_y\left(\lim_{T\rightarrow\infty}\mathcal C_{0,T}\mid x_0=x,\mu\right),
\]
we immediately show that $V^*(x,y) =V_{u^*}(x,y)$. By the fixed point equation $\mathbf T[V^*](x,y)=V^*(x,y)$ and $\mathbf T_{u^*}[V_{u^*}](x,y)=V_{u^*}(x,y)$, this further implies \eqref{eq:opt_eq} holds.

Second suppose $u^*$ satisfies the equality in \eqref{eq:opt_eq}. Then by the fixed point equality $\mathbf T[V^*](x,y)=V^*(x,y)$, we immediately obtain the equation $V^*(x,y)=\mathbf T_{u^*}[V^*](x,y)$ for any $x\in\mathcal X$ and $y\in\mathcal Y$. since the fixed point solution to $\mathbf T_{u^*}[V](x,y)=V(x,y)$ is unique, we further show that $\mathbf T[V^*](x,y)=V^*(x,y)=V_{u^*}(x,y)$ and $V_{u^*}(x,y)=\min_{\mu\in\Pi_H} \text{CVaR}_y\left(\lim_{T\rightarrow\infty}\mathcal C_{0,T}\mid x_0=x,\mu\right)$ from Theorem \ref{thm:MDP_CVaR}. By using the policy construction formula in \eqref{eq:policy_construct} to obtain the history dependent policy $u^*_H$ and following the above arguments at which the augmented Markovian stationary policy $u$ is replaced by $u^*$, this further implies
\[
\min_{\mu\in\Pi_H} \text{CVaR}_y\left(\lim_{T\rightarrow\infty}\mathcal C_{0,T}\mid x_0=x,\mu\right)=\text{CVaR}_y\left(\lim_{T\rightarrow\infty}\mathcal C_{0,T}\mid x_0=x,u^*_H\right),
\]
 i.e., $u^*$ is an optimal augmented stationary Markovian policy.

\subsection{Proof of Lemma \ref{lem:Bellman_prop_inter}}
We first proof the monotonicity property. Based on the definition of $\interpI_{x}[V](y)$, if $V_1(x,y)\geq V_2(x,y)$ $\forall x\in\mathcal X$ and $y\in\mathcal Y$, we have that
\[
\interpI_{x}[V_1](y)=\frac{y_{i+1}V_1(x,y_{i+1})(y-y_i)+y_iV_1(x,y_{i})(y_{i+1}-y)}{y_{i+1}-y_i},\,\text{if $y\in\mathbf I_i(x)$}.
\]
Since $y_i,y_{i+1}\in\mathcal Y$ and $(y_{i+1}-y),(y-y_i)\geq 0$ (because $y\in\mathbf I_i(x)$), we can easily see that $\interpI_{x}[V_1](y)\geq \interpI_{x}[V_2](y)$. As $y\in\mathcal Y$ and $\xi(\cdot)P(\cdot|x,a)\geq 0$ for any $\xi\in \U_{\text{CVaR}}(y, P(\cdot|x,a)$, this further implies $\bellint[V_1](x,y)\geq \bellint[V_2](x,y)$.

Next we prove the constant shift property. Note from the definition of $\interpI_{x}[V](y)$ that
\[
\begin{split}
&\interpI_{x}[V+K](y)\\
=&y_i(V(x,y_{i})+K)+\frac{y_{i+1}(V(x,y_{i+1})+K)-y_i(V(x,y_{i})+K)}{y_{i+1}-y_i}(y-y_i),\,\text{if $y\in\mathbf I_i(x)$},\\
=&yK+y_iV(x,y_{i})+\frac{y_{i+1}V(x,y_{i+1})-y_iV(x,y_{i})}{y_{i+1}-y_i}(y-y_i),\,\text{if $y\in\mathbf I_i(x)$}\\
=&\interpI_{x}[V](y)+yK.
\end{split}
\]
Therefore by definition of $\bellint[V](x,y)$, the constant shift property: $\bellint[V+K](x,y)=\bellint[V](x,y)+\gamma K$ for any $x\in\mathcal X$, $y\in\mathcal Y$, follows directly from the above arguments.

Equipped with both properties in monotonicity and constant shift, the proof of contraction of $\bellint$ directly follows from the analogous proof in Lemma \ref{lem:prop_Bellman}.

Finally we prove the concavity preserving property.
Assume $yV(x,y)$ is concave in $y\in\mathcal Y$ for any $x\in\mathcal X$. Then for $y_{i+2}>y_{i+1}>y_i$, $\forall i\in\{1,\ldots,N(x)-2\}$ the following inequality immediately follows from the definition of a concave function:
\begin{equation}\label{ineq:concave}
\begin{split}
\frac{d \interpI_{x}[V](y)}{dy}&\bigg\vert_{y\in \mathbf I_{i+1}(x)}=\frac{y_{i+1}V(x,y_{i+1})-y_iV(x,y_i)}{y_{i+1}-y_i}\\
\geq& \frac{y_{i+2}V(x,y_{i+2})-y_{i+1}V(x,y_{i+1})}{y_{i+2}-y_{i+1}}=\frac{d \interpI_{x}[V](y)}{dy}\bigg\vert_{y\in \mathbf I_{i+2}(x)}.
\end{split}
\end{equation}
We then show that the following inequality in each of the following cases, whenever the slope exists:
\[
\interpI_{x}[V](z_1)\le  \interpI_{x}[V](z_2)+\frac{d \interpI_{x}[V](y)}{dy}\bigg\vert_{y=z_2}(z_1-z_2),\,\,\forall z_1,z_2\in\mathcal Y\setminus\{0\}.
\]
\noindent (1) There exists $i\in\{1,\ldots,N(x)-1\}$ such that $z_1,z_2\in\mathbf I_{i+1}(x)$. In this case we have that
\[
\frac{d \interpI_{x}[V](y)}{dy}\bigg\vert_{y=z_1}=\frac{d \interpI_{x}[V](y)}{dy}\bigg\vert_{y=z_2},
\]
and this further implies
\[
\interpI_{x}[V](z_1)=  \interpI_{x}[V](z_2)+\frac{d \interpI_{x}[V](y)}{dy}\bigg\vert_{y=z_2}(z_1-z_2).
\]
\noindent (2) There exists $i,j\in\{1,\ldots,N(x)-2\}$, $i+1<j$ such that $z_1\in\mathbf I_{i+1}(x)$ and $z_2\in\mathbf I_j(x)$. In this case, without loss of generality we assume $j=i+1$. The proof for case: $j>i+2$ is omitted for the sake of brevity, as it can be completed by iteratively applying the same arguments from case: $j=i+2$. Since $z_1\in\mathbf I_i(x)$, $z_2\in\mathbf I_j(x)$, we have $z_2-z_1\ge 0$ and
\[
 \frac{d \interpI_{x}[V](y)}{dy}\bigg\vert_{y=z_1}\geq\frac{d \interpI_{x}[V](y)}{dy}\bigg\vert_{y=z_2}.
\]
Based on the definition of the linear interpolation function, we have that
\[
\interpI_{x}[V](y_{i+1})=y_{i+1}V(x,y_{i+1})= \interpI_{x}[V](y_{i})+\frac{d \interpI_{x}[V](y)}{dy}\bigg\vert_{y\in\mathbf I_{i+1}(x)}(y_{i+1}-y_{i}).
\]
Furthermore, combining previous arguments with the definitions of $\interpI_{x}[V](z_1)$, $\interpI_{x}[V](z_2)$ implies that for $(z_2-y_{i+1})\geq 0$,
\[
\begin{split}
\interpI_{x}[V](z_2)=&\interpI_{x}[V](y_{i+1})+\frac{d \interpI_{x}[V](y)}{dy}\bigg\vert_{y=z_2}(z_2-y_{i+1})\\
\leq &\interpI_{x}[V](y_{i+1})+\frac{d \interpI_{x}[V](y)}{dy}\bigg\vert_{y=z_1}(z_2-y_{i+1})\\
=&\interpI_{x}[V](y_{i})+\frac{d \interpI_{x}[V](y)}{dy}\bigg\vert_{y\in\mathbf I_{i+1}(x)}(z_2-y_{i})\\
=&\interpI_{x}[V](z_1)+\frac{d \interpI_{x}[V](y)}{dy}\bigg\vert_{y=z_1}(z_2-z_1).
\end{split}
\]
\noindent (3) There exists $i,j\in\{1,\ldots,N(x)-2\}$, $i+1<j$ such that $z_2\in\mathbf I_{i+1}(x)$ and $z_1\in\mathbf I_j(x)$. In this case, without loss of generality we assume $j=i+1$. The proof for case: $j>i+2$ is omitted for the sake of brevity, as it can be completed by iteratively applying the same arguments from case: $j=i+2$. Since $z_2\in\mathbf I_{i+1}(x)$, $z_1\in\mathbf I_j(x)$, we have $z_1-z_2\ge 0$ and
\[
\frac{d \interpI_{x}[V](y)}{dy}\bigg\vert_{y=z_1}\leq\frac{d \interpI_{x}[V](y)}{dy}\bigg\vert_{y=z_2}.
\]
Similar to the analysis in the previous case, we have that
\[
\interpI_{x}[V](y_{i})=y_{i}V(x,y_{i})= \interpI_{x}[V](y_{i+1})+\frac{d \interpI_{x}[V](y)}{dy}\bigg\vert_{y\in\mathbf I_{i+1}(x)}(y_{i}-y_{i+1})\\
\]
Furthermore, combining previous arguments with the definitions of $\interpI_{x}[V](z_1)$, $\interpI_{x}[V](z_2)$ implies that for $(z_2-z_1)\leq 0$,
\[
\begin{split}
\interpI_{x}[V](z_2)=&\interpI_{x}[V](y_{i})+\frac{d \interpI_{x}[V](y)}{dy}\bigg\vert_{y=z_2}(z_2-y_{i})\\
= &\interpI_{x}[V](y_{i+1})+\frac{d \interpI_{x}[V](y)}{dy}\bigg\vert_{y=z_2}(z_2-y_{i+1})\\
= &\interpI_{x}[V](z_1)+\frac{d \interpI_{x}[V](y)}{dy}\bigg\vert_{y=z_2}(z_2-z_1)\\
\leq&\interpI_{x}[V](z_1)+\frac{d \interpI_{x}[V](y)}{dy}\bigg\vert_{y=z_1}(z_2-z_1).
\end{split}
\]

Thus we have just shown that the first order sufficient condition for concave functions, corresponding to ${\interpI_{x}[V](y)}$, holds, i.e., ${\interpI_{x}[V](y)}$ is concave in $y\in\mathcal Y\setminus\{0\}$ for any given $x\in\mathcal X$.  Now since ${\interpI_{x}[V](y)}$ is a continuous piecewise linear function in $y\in\mathcal Y$ and a concave function when the domain is restricted to $\mathcal Y\setminus\{0\}$. By continuity this immediately implies that  ${\interpI_{x}[V](y)}$ is concave in $y\in\mathcal Y$ as well. Then following the identical arguments in the proof of Lemma \ref{lem:prop_Bellman} for the concavity preserving property, we can thereby show that
\[
y\bellint[V](x,y) =\min_{a\in\mathcal A}\left\{yC(x,a)+\max_{\xi\in \U_{\text{CVaR}}(y, P(\cdot|x,a))}\sum_{x'\in\mathcal X}{\interpI_{x'}[V](y\xi(x'))}P(x'|x,a)\right\}
\]
is concave in $y\in\mathcal Y$ for any given $x\in\mathcal X$.

\subsection{Useful Intermediate Results}
\begin{lemma}\label{lem:Interp_is_Lipschitz_preserving}
Let $f(y):[0,1]\to \R$ be a concave function, differentiable almost everywhere, with Lipschitz constant $M$. Then the linear interpolation $\interpI [f](y)$ is also concave, and with Lipschitz constant $M_I \leq M$.
\end{lemma}

\paragraph{Proof}
For every segment $[y_j,y_{j+1}]$ in the linear interpolation, $f(y)$ is concave, and with Lipschitz constant $M$, and $\interpI [f](y)$ is linear. Also, $f(y_j) = \interpI [f](y_j)$, and $f(y_{j+1}) = \interpI [f](y_{j+1})$, by definition of the linear interpolation. Denote by $c_{j} $ the magnitude of the slope of $\interpI [f](y)$ at $y\in[y_j,y_{j+1}]$.

Assume by contradiction that $c_j > \max_{y\in[y_j,y_{j+1}]}|f'(y)|$ whenever $f'(y)$ exists.
 Consider the case when $f(y_{j+1}) \geq f(y_{j})$. This implies $c_{j}$ is the slope of the interpolation function $\interpI [f](y)$ at $y\in[y_j,y_{j+1}]$.  Then by the fundamental theorem of calculus, we have
\begin{equation*}
    f(y_{j+1}) - f(y_{j})= \int_{y_j}^{y_{j+1}} f'(y)dy\leq \int_{y_j}^{y_{j+1}}|f'(y)|dy < \int_{y_j}^{y_{j+1}} c_j dy = (\interpI [f](y_{j+1}) - \interpI [f](y_{j})),
\end{equation*}
contradicting $f(y_{j+1}) = \interpI [f](y_{j+1})$ and $f(y_j) = \interpI [f](y_j)$.

On the other hand, consider the case when $f(y_{j+1}) \leq f(y_{j})$.  This implies $-c_{j}$ is the slope of the interpolation function $\interpI [f](y)$ at $y\in[y_j,y_{j+1}]$. Again by  fundamental theorem of calculus,
\begin{equation*}
   0\leq f(y_{j+1}) - f(y_{j})= \int_{y_j}^{y_{j+1}} f'(y)dy\geq \int_{y_j}^{y_{j+1}}-|f'(y)|dy > \int_{y_j}^{y_{j+1}} -c_j dy = \interpI [f](y_{j}) - \interpI [f](y_{j+1}).
\end{equation*}
Since $f(y_{j+1}) = \interpI [f](y_{j+1})$ and $f(y_j) = \interpI [f](y_j)$, which implies $\interpI [f](y_{j})-\interpI [f](y_{j+1})\geq 0$, the above expression clearly leads to a contradiction.

We finally have that $ \max_{y\in[y_j,y_{j+1}]} |f'(y)| \geq c_j $ for segment $j\in\{1,\ldots,N(x)-1\}$. As this argument holds for each segment, by maximizing over $j$ over $\{1,\ldots,N(x)-1\}$, we have that
\[
M \geq \max_{j\in\{1,\ldots,N(x)-1\}}\max_{y\in[y_j,y_{j+1}]} |f'(y)| \geq \max_{j\in\{1,\ldots,N(x)-1\}}c_j = M_I .
\]

The concavity property (thus differentiability almost everywhere) are well-known results of linear interpolation \cite{phillips2003interpolation}.

\begin{lemma}\label{lem:Lipschitz_Bellman}
Let $yV(x,y)$ be Lipschitz with constant $M$, concave, and differentiable almost everywhere, for every $x\in \mathcal X$ and $y\in [0,1]$. Then $y\mathbf T[V](x,y)$ is also Lipschitz with constant $C_{\max} + \gamma M$.
\end{lemma}

\paragraph{Proof}
For any given state-action pair $x\in \mathcal X$, and $a\in \mathcal A$, let $P(x') = P(x'|x,a)$ be the transition kernel. Consider the function
\begin{equation*}
H(y) \doteq \max_{\xi\in \U_{\text{CVaR}}(y, P(\cdot))}\sum_{x'\in\mathcal X} y \xi(x')V\left(x',y\xi(x')\right)P(x').
\end{equation*}
Note that, by definition of $\U_{\text{CVaR}}$, and a change of variables $z(x') = y \xi(x')$, we can write $H(y)$ as follows:
\begin{equation}\label{eq:H_function_definition}
H(y) = \max_{\substack{0\leq z(x')\leq 1,\\ \sum_{x'} P(x')z(x') = y}}\sum_{x'\in\mathcal X} z(x')V\left(x',z(x')\right)P(x').
\end{equation}
The Lagrangian of the above maximization problem is
\begin{equation*}
    L(z,\lambda;y) = \sum_{x'\in\mathcal X} z(x')V\left(x',z(x')\right)P(x') -\lambda (\sum_{x'} P(x')z(x') - y).
\end{equation*}
Since $yV(x,y)$ is concave, the maximum is attained. By first order optimality condition the following expression holds:
\begin{equation*}
    \frac{\partial L(z,\lambda;y)}{\partial z(x')} = P(x') \frac{\partial \left[ z(x')V\left(x',z(x')\right)\right]}{\partial z(x')} -\lambda P(x') = 0.
\end{equation*}
Summing the last expression over $x'$, we obtain:
\begin{equation*}
    \sum_{x'\in\mathcal X} P(x') \frac{\partial \left[ z(x')V\left(x',z(x')\right)\right]}{\partial z(x')} = \sum_{x'\in\mathcal X} \lambda P(x') = \lambda.
\end{equation*}
Now, from the Lipschitz property of $yV(x,y)$, we have
\begin{equation*}
    \left|\sum_{x'\in\mathcal X} \lambda P(x')\right|\leq \sum_{x'\in\mathcal X} P(x') \left| \frac{\partial \left[ z(x')V\left(x',z(x')\right)\right]}{\partial z(x')} \right| \leq \sum_{x'\in\mathcal X} P(x') M = M.
\end{equation*}
Thus,
\begin{equation*}
    |\lambda| \leq \sum_{x'\in\mathcal X} P(x') \left| \frac{\partial \left[ z(x')V\left(x',z(x')\right)\right]}{\partial z(x')} \right| \leq M.
\end{equation*}
Note that the objective in \eqref{eq:H_function_definition} does not depend on $y$. From the envelope theorem \cite{milgrom2002envelope}, it follows that
\begin{equation*}
    \frac{d H(y)}{d y} = \lambda,
\end{equation*}
therefore, $H(y)$ is Lipschitz, with constant $M$.

Now, by definition,
\begin{equation*}
    y\mathbf T[V](x,y) = \min_{a\in\mathcal A}\left[y C(x,a)+\gamma\max_{\xi\in \U_{\text{CVaR}}(y, P(\cdot|x,a))}\sum_{x'\in\mathcal X}y \xi(x')V\left(x',y\xi(x')\right)P(x'|x,a)\right].
\end{equation*}
Using our Lipschitz result for $H(y)$, we have that for any $a \in \mathcal A$, the function
\begin{equation*}
    y C(x,a)+\gamma\max_{\xi\in \U_{\text{CVaR}}(y, P(\cdot|x,a))}\sum_{x'\in\mathcal X}y \xi(x')V\left(x',y\xi(x')\right)P(x'|x,a)
\end{equation*}
is Lipschitz in $y$, with constant $C(x,a)+\gamma M$. Using again the envelope theorem \cite{milgrom2002envelope}, we obtain that $y\mathbf T[V](x,y)$ is Lipschitz, with constant $C_{\max}+\gamma M$.

\begin{lemma}\label{lem:Lipschitz_bound}
Consider Algorithm \ref{alg:CVI}. Assume that for any $x\in \mathcal X$, the initial value function satisfies that $yV_0(x,y)$ is Lipschitz (in $y$), with uniform constant $M_0$. We have that for any $t\in\{0,1,\ldots,\}$, the function $yV_{t}(x,y)$ is Lipschitz in $y$ for any $x\in\mathcal X$, with Lipschitz constant
\[
M_{t}=\frac{1-\gamma^t}{1-\gamma}C_{\max}+\gamma^t M_0\leq \frac{C_{\max}}{1-\gamma}+M_0,\,\,\forall t.
\]
\end{lemma}

\paragraph{Proof}
Let $\bellint[V]$ denote the application of the Bellman operator $\mathbf T$ to the linearly-interpolated version of $y V(x,y)$. We have, by definition, that
\begin{equation*}
    V_{1}(x,y) = \bellint[V_0](x,y).
\end{equation*}
Using Lemma \ref{lem:Interp_is_Lipschitz_preserving} and Lemma \ref{lem:Lipschitz_Bellman}, we have that $V_{1}(x,y)$ is Lipschitz, with $M_{1}\leq C_{\max} + \gamma M_0$.

Note now, that $V_{2}(x,y) = \bellint [V_1](x,y)$. Thus, by induction, we have
\begin{equation*}
    M_{t} \leq \frac{1-\gamma^t}{1-\gamma}C_{\max}+\gamma^t M_0,
\end{equation*}
and the result follows.

\subsection{Proof of Theorem \ref{thm:interpolation_bdd}}
The proof of this theorem is split into three parts. In the first part, we bound the difference  ${\interpI_{x}[V_{t}](y)}/{y}-V_{t}(x,y)$ at each state $(x,y)\in\mathcal X\times\mathcal Y$ using the previous technical lemmas and Lipschitz property.

In the second part, we bound the difference of $\bellint[V_t](x,y)- {\mathbf T}[V_t](x,y)$.

In the third part we bound the interpolation error using contraction properties of Bellman recursions.

First we analyze the bounds for ${\interpI_{x}[V_{t}](y)}/{y}-V_{t}(x,y)$ in the following four cases. Notice that from Lemma \ref{lem:Lipschitz_bound}, we have that $|{d \interpI_{x}[V_t](y)}/{dy}|\leq M:={C_{\max}}/{(1-\gamma)}+M_0$.

\noindent (1) When $y=0$ (for which $y\in\mathbf I_1(x)$). \\
Using previous analysis and L'Hospital's rule we have that $\lim_{y\rightarrow 0}{\interpI_{x}[V_t](y)}/{y}=V_t(x,0)$.
This further implies $\lim_{y\rightarrow 0}{\interpI_{x}[V_t](y)}/{y} - V_t(x,0)=0$.

\noindent (2) When $y\in\mathbf I_{i+1}(x)$, $2\leq i< N(x)-1$. \\
Similar to the inequality in \eqref{ineq:concave}, by concavity of $yV_t(x,y)$ in $y\in\mathcal Y$, we have that
\[
\frac{d \interpI_{x}[V_t](y)}{dy}\bigg\vert_{y\in\mathbf I_{i+1}(x)}=\frac{y_{i+1}V_t(x,y_{i+1})-y_{i}V_t(x,y_{i})}{y_{i+1}-y_{i}}\leq \frac{yV_t(x,y)-y_{i}V_t(x,y_{i})}{y-y_{i}},
\]
and
\[
\frac{d \interpI_{x}[V_t](y)}{dy}\bigg\vert_{y\in\mathbf I_{i+2}(x)}=\frac{y_{i+2}V_t(x,y_{i+2})-y_{i+1}V_t(x,y_{i+1})}{y_{i+2}-y_{i+1}}\leq \frac{y_{i+1}V_t(x,y_{i+1})-yV_t(x,y)}{y_{i+1}-y}.
\]
From the first inequality, for each $(x,y)\in\mathcal X\times\mathcal Y$ we get,
\begin{equation}\label{eq:upper_bdd}
\frac{\interpI_{x}[V_t](y)}{y} - V_t(x,y)\leq \frac{1}{y}\left(y_{i}V_t(x,y_{i})+\frac{y_{i+1}V_t(x,y_{i+1})-y_{i}V_t(x,y_{i})}{y_{i+1}-y_{i}}(y-y_i)-yV_t(x,y)\right)\leq 0.
\end{equation}

On the other hand, rearranging the second inequality gives
\begin{equation}\label{eq:small_bdd}
\begin{split}
&\frac{1}{y}(\interpI_{x}[V_t](y) - yV_t(x,y))\\
\geq&\frac{1}{y}\left(y_iV_t(x,y_{i})+\frac{d \interpI_{x}[V_t](y)}{dy}\bigg\vert_{y\in\mathbf I_{i+1}(x)}(y-y_i)-y_{i+1}V_t(x,y_{i+1})-\frac{d \interpI_{x}[V_t](y)}{dy}\bigg\vert_{y\in\mathbf I_{i+2}(x)}(y-y_{i+1})\right)\\
= &\left(\frac{d \interpI_{x}[V_t](y)}{dy}\bigg\vert_{y\in\mathbf I_{i+1}(x)}-\frac{d \interpI_{x}[V_t](y)}{dy}\bigg\vert_{y\in\mathbf I_{i+2}(x)}\right)\frac{y-y_{i+1}}{y}\geq -2M\left(\frac{y_{i+1}}{y}-1\right).
\end{split}
\end{equation}
Furthermore by the Lipschitz property, we also have the following inequality as well:
\begin{equation}\label{eq:large_bdd}
\begin{split}
&\frac{1}{y}(\interpI_{x}[V_t](y) - yV_t(x,y))\\
=&\frac{y_{i+1}V_t(x,y_{i+1})(y-y_i)+y_iV_t(x,y_{i})(y_{i+1}-y)}{(y_{i+1}-y_i)y}-V_t(x,y)\\
\geq &\frac{y_{i}V_t(x,y_{i})(y-y_i)+y_iV_t(x,y_{i})(y_{i+1}-y)-M(y_{i+1}-y_i)(y-y_i)}{(y_{i+1}-y_i)y}-V_t(x,y)\\
= &\frac{y_{i}V_t(x,y_{i})-M(y-y_i)}{y}-V_t(x,y)\geq -2M\left(1-\frac{y_{i}}{y}\right).
\end{split}
\end{equation}
Combining the inequalities \eqref{eq:small_bdd} and \eqref{eq:large_bdd}, the following lower bound for ${\interpI_{x}[V_{t}](y)}/{y}-V_{t}(x,y)$ holds:
\[
\frac{1}{y}(\interpI_{x}[V_t](y) - yV_t(x,y))\geq \delta:=-2M\min\left\{1-\frac{y_{i}}{y},\frac{y_{i+1}}{y}-1\right\},\,\forall y\in\mathbf I_{i+1}(x),\,i\geq 2.
\]
From the above definition, when $y_i\leq y\leq (y_i+y_{i+1})/2$, the lower bound becomes $\delta = -2M(1-{y_{i}}/{y})$ and when $(y_i+y_{i+1})/2\leq y\leq y_{i+1}$, the corresponding lower bound is $\delta = -2M({y_{i+1}}/{y}-1)$. In both cases, $\delta$ is minimized when $y=(y_i+y_{i+1})/2$. Therefore, the above analysis implies the following lower bound:
\[
\frac{1}{y}(\interpI_{x}[V_t](y) - yV_t(x,y))\geq-2M\frac{y_{i+1}-y_{i}}{y_{i+1}+y_{i}},\,\forall y\in\mathbf I_{i+1}(x),\,i\geq 2.
\]
When $y_{i+1}=\theta y_i$ for $i\in\{2,\ldots,N(x)-1\}$ for some constant $\theta\geq 1$, this further implies that
\[
\frac{1}{y}(\interpI_{x}[V_t](y) - yV_t(x,y))\geq-2M\frac{\theta-1}{\theta+1}\geq -M(\theta-1),\,\,\forall y\in\mathcal Y\setminus[0,\epsilon].
\]
Then combining the results, here we get the following bound for ${\interpI_{x}[V_{t}](y)}/{y}-V_{t}(x,y)$:
\[
-M(\theta-1)\leq\frac{\interpI_{x}[V_t](y)}{y} - V_t(x,y)\leq 0,\,\forall y\in\mathbf I_{i+1}(x),\,i\geq 2.
\]
\noindent (3) When $y\in\mathbf I_{N(x)}(x)$, i.e., $y\in(y_{N(x)-1},1]$. \\
Similar to the proof of case (2), we can show that for any $x\in\mathcal X$ and $y\in\mathbf I_{N(x)}(x)$, the same lines of arguments in inequality \eqref{eq:upper_bdd} and \eqref{eq:large_bdd} hold, which implies
\[
-2M\left(1-y_{N(x)-1}\right)\leq-2M\left(1-\frac{y_{N(x)-1}}{y}\right)\leq\frac{1}{y}(\interpI_{x}[V_t](y) - yV_t(x,y))\leq 0.
\]
When $y_{N(x)}=1=\theta y_{N(x)-1}$, this further shows that
\[
-2My_{N(x)-1}(\theta-1)=-2M\left(y_{N(x)}-y_{N(x)-1}\right)\leq\frac{1}{y}(\interpI_{x}[V_t](y) - yV_t(x,y))\leq 0,
\]
and
\[
-2M(\theta-1)\leq-\frac{2}{\theta}M(\theta-1)\leq\frac{1}{y}(\interpI_{x}[V_t](y) - yV_t(x,y))\leq 0.
\]
\noindent (4) When $y\in\mathbf I_2(x)$, i.e., $y\in(0,y_2]$. \\
From inequality \eqref{eq:upper_bdd}, the definition of $\interpI_{x}[V_t](y)$, we have that
\[
0\geq \frac{\interpI_{x}[V_t](y) - yV_t(x,y)}{y} = \frac{y(V_t(x,y_2)-V_t(x,y))}{y}=V_t(x,y_2)-V_t(x,y)\geq V_t(x,y_2)-V_t(x,0).
\]
The first inequality is due to the fact that $yV_t(x,y)$ is concave in $y\in\mathcal Y$ for any $x\in\mathcal X$, thus the first order condition implies
\[
\frac{y_{2}V_n(x,y_{2})-y_{1}V_n(x,y_{1})}{y_{2}-y_{1}}\leq \frac{yV_n(x,y)-y_{1}V_n(x,y_{1})}{y-y_{1}},\,\,\forall y\in\mathbf I_{2}(x),
\]
and the last inequality is due to the similar fact that
\[
V_t(x,w)=\frac{wV_t(x,w)-0\cdot V_t(x,0)}{w-0}\leq \frac{zV_t(x,z)-0\cdot V_t(x,0)}{z-0}=V_t(x,z),\,\,\forall z,w\in\mathcal Y, \,\,z\leq w.
\]
Therefore the condition of this theorem implies
\[
0\geq \frac{\interpI_{x}[V_t](y) - yV_t(x,y)}{y} \geq -\epsilon,\,\,\forall t\geq 0,\,x\in\mathcal X,\,\,y\in\mathcal Y.
\]

Combining the above four cases, we have that for each state $(x,y)\in\mathcal X\times\mathcal Y$,
\[
0\geq \frac{\interpI_{x}[V_{t}](y)}{y}-V_{t}(x,y)\geq -2M(\theta-1)-\epsilon,\,\,\forall t.
\]

Second, we bound the difference of $\bellint[V_t](x,y)- {\mathbf T}[V_t](x,y)$. By recalling that $\xi(\cdot)P(\cdot|x,a)$ is a probability distribution for any $\xi\in \U_{\text{CVaR}}(y, P(\cdot|x,a))$, we then combine all previous arguments and show that at any $t\in\{0,1,\ldots,\}$ and any $x\in\mathcal X$, $a\in\mathcal A$, $y\in\interpY(x)$,
\[
\max_{\xi\in \U_{\text{CVaR}}(y, P(\cdot|x,a))}\sum_{x'\in\mathcal X,\xi(x')\neq 0}\left(\frac{\interpI_{x'}[V_{t}](y\xi(x'))}{y\xi(x')}-V_{t}(x',y\xi(x'))\right)\xi(x')P(x'|x,a)\geq-2M(\theta-1)-\epsilon.
\]
This further implies
\begin{equation}\label{eq:bounds}
 {\mathbf T}[V_t](x,y)-\gamma (2M(\theta-1)+\epsilon)\leq\bellint[V_t](x,y)\leq {\mathbf T}[V_t](x,y).
\end{equation}

Third, we prove the error bound of interpolation based value iteration using the above properties. By putting $t=0$ in \eqref{eq:bounds}, we have that
\[
-\gamma (2M(\theta-1)+\epsilon)\leq\bellint[V_0](x,y)-{\mathbf T}[V_0](x,y)\leq 0.
\]
Applying the Bellman operator $\mathbf T$ on all sides of the above inequality and noting that $\mathbf T$ is a translational invariant mapping, the above expression implies
\[
 {\mathbf T}^2[V_0](x,y)-\gamma^2 (2M(\theta-1)+\epsilon)\leq\mathbf T[\bellint[V_0]](x,y)=\mathbf T[{V}_1](x,y)\leq {\mathbf T}^2[V_0](x,y).
\]
By adding the inequality: $- \gamma (2M(\theta-1)+\epsilon)\leq\bellint[V_1](x,y)-{\mathbf T}[V_1](x,y)\leq 0$
to the above expression, this further implies the following expression:
\[
 {\mathbf T}^2[V_0](x,y)-\gamma(1+\gamma)(2M(\theta-1)+\epsilon)\leq\bellint[V_1](x,y)=\bellint^2[V_0](x,y)\leq {\mathbf T}^2[V_0](x,y).
 \]
 Then, by repeating this process, we can show that for any $n\in\natural$, the following inequality holds:
 \[
 {\mathbf T}^n[V_0](x,y)-\gamma\frac{1-\gamma^{n}}{1-\gamma}(2M(\theta-1)+\epsilon)\leq\bellint^n[V_0](x,y)\leq {\mathbf T}^n[V_0](x,y).
 \]
Note that when $n\rightarrow\infty$,  we have that $\gamma^n$ converges to $0$, ${\mathbf T}^n[V_0](x,y)$ converges to $\min_{\mu\in\Pi_H} \text{CVaR}_{y}\left(\lim_{T\rightarrow\infty}\mathcal C_{0,T}\mid x,\mu\right)$ (follow from Theorem \ref{thm:MDP_CVaR}) and $\bellint^n[V_0](x,y)$ converges to $\widehat{V}^*(x,y)$ (follow from the contraction property in Lemma \ref{lem:Bellman_prop_inter}).

Furthermore, from Proposition 1.6.4 in \cite{Ber2012DynamicProgramming}, the contraction property of Bellman operator $\mathbf T$ implies that for any $x\in\mathcal X$, $y\in\mathcal Y$, the following expression holds:
\[
|{\mathbf T}^n[V_0](x,y)-V^*(x,y)|\leq\frac{\gamma^n}{1-\gamma}(C_{\max}+\|Z\|_\infty)
\]
where $Z$ is the bounded random variable of the initial value function $V_0(x,y)=\text{CVaR}_y(Z\mid x_0=x)$ such that $\|V_0\|_\infty\leq\|Z\|_\infty$, and $V^*(x,y)=\min_{\mu\in\Pi_H} \text{CVaR}_{y}\left(\lim_{T\rightarrow\infty}\mathcal C_{0,T}\mid x,\mu\right)$. This further implies for any $x\in\mathcal X$, $y\in\mathcal Y$,
\[
|\bellint^n[V_0](x,y)-V^*(x,y)|\leq \gamma\frac{1-\gamma^{n}}{1-\gamma}(2M(\theta-1)+\epsilon)+\frac{\gamma^n}{1-\gamma}(C_{\max}+\|Z\|_\infty).
\]
Then, by combining all the above arguments, we prove the claim of this theorem.

\section{Trajectory Plots}
\begin{figure}
\begin{center}
  \includegraphics[width=0.45\textwidth]{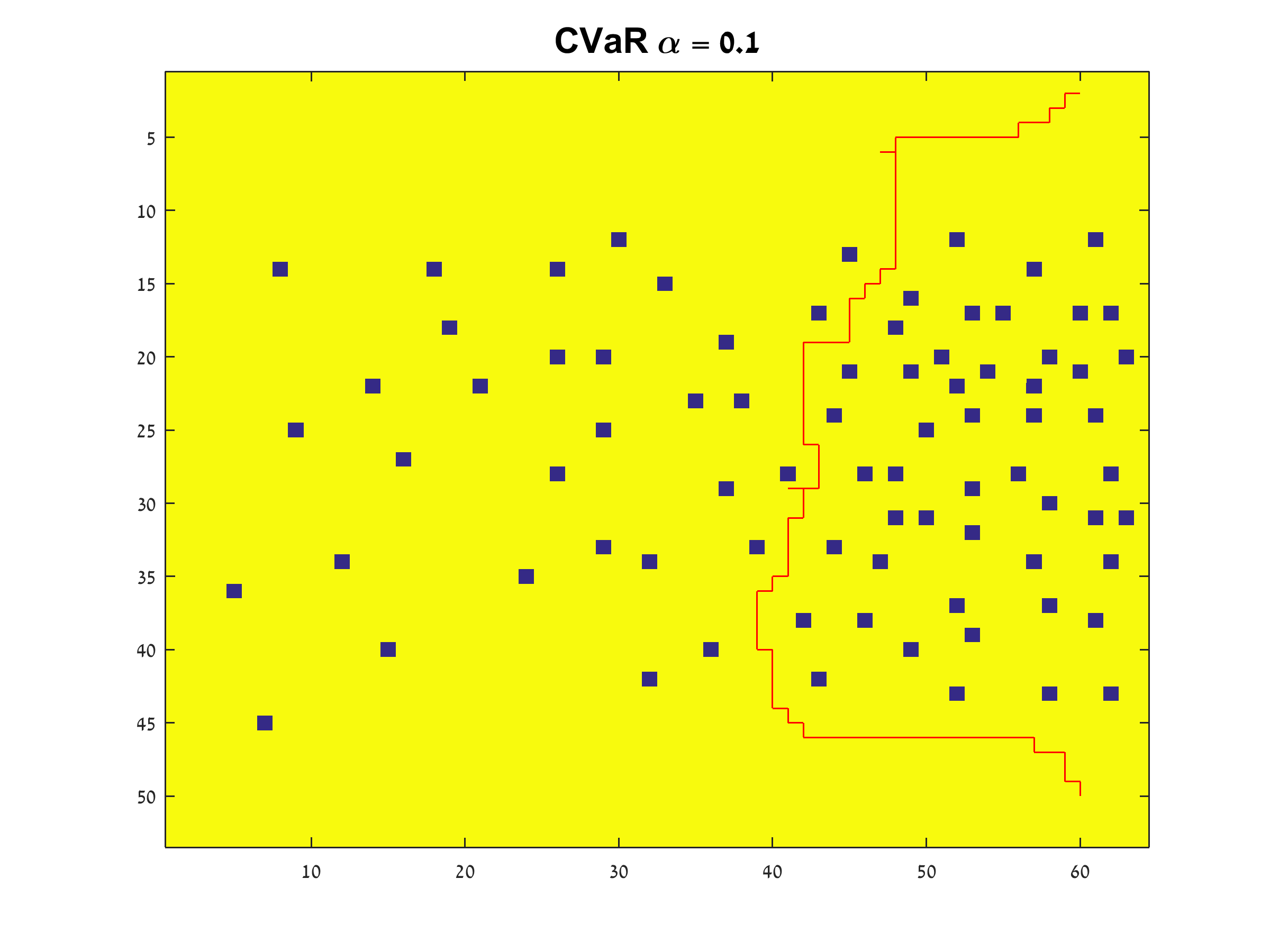}
  \includegraphics[width=0.45\textwidth]{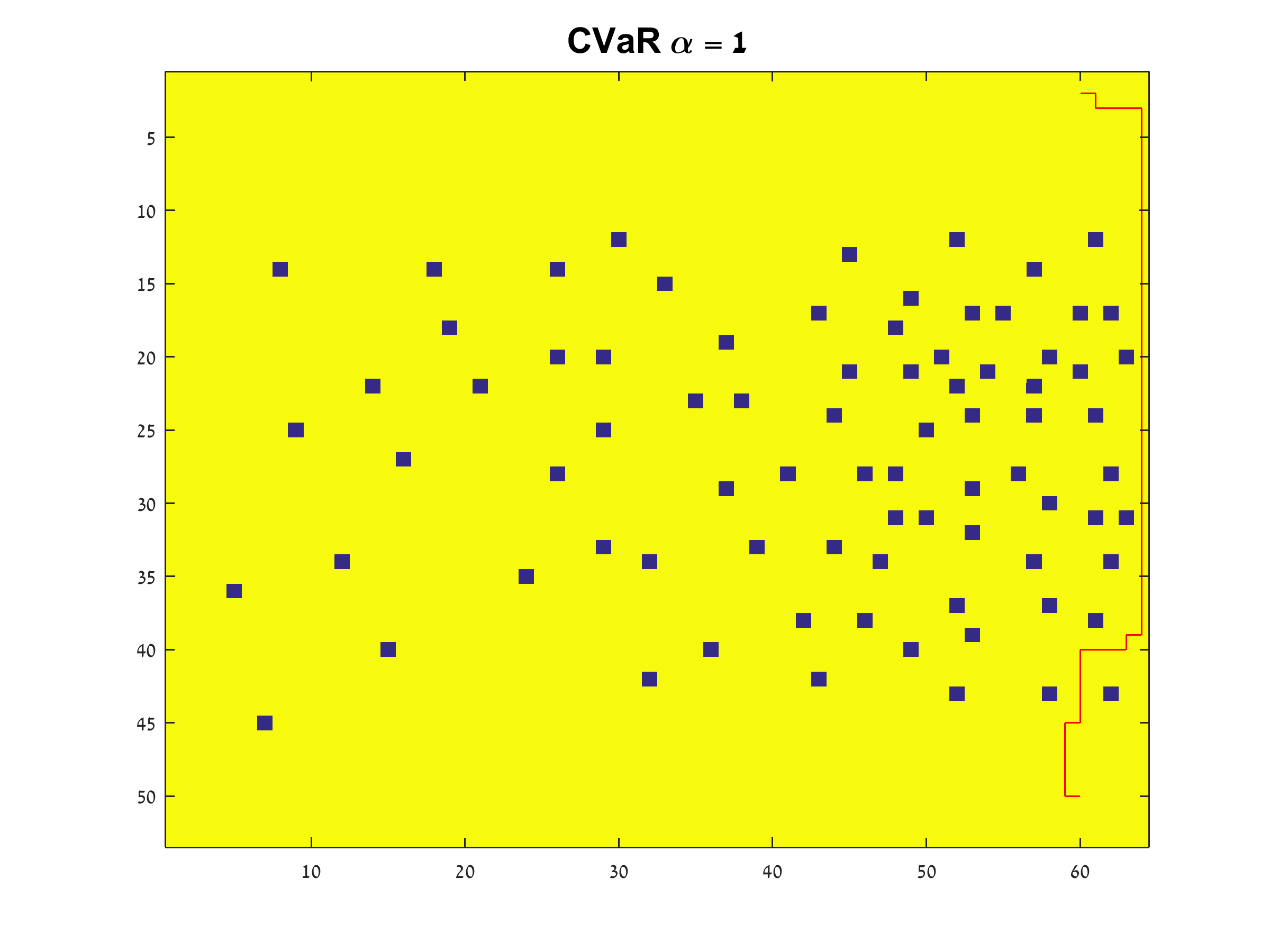}
  \caption{Grid-world - trajectory plots.}\label{fig:gridworld_traj}
\end{center}
\end{figure}

In Figure \ref{fig:gridworld_traj} we demonstrate simulated trajectories according to a policy that is greedy w.r.t. the value function, according to Theorem \ref{thm:policy}.

\section{Generalization to Mean-CVaR Optimization}\label{subsec:mean-cvar}

In this section we extend our approach to MDPs with a mean-CVaR objective of the form:
\begin{equation}\label{eq:mean-CVaR}
\min_{\mu\in\Pi_H} \quad \lambda\mathbb E\left(\lim_{T\rightarrow\infty}\mathcal C_{0,T}\mid x_0,\mu\right) +(1-\lambda) \text{CVaR}_\alpha\left(\lim_{T\rightarrow\infty}\mathcal C_{0,T}\mid x_0,\mu\right),
\end{equation}
where $\lambda \in [0,1]$. Such an objective is common in practice \cite{iyengar2013fast}, and is also useful for solving CVaR-constrained objectives using standard Lagrangian methods (see, e.g.,~ \cite{chow2014cvar}).

Now for any $\alpha_1,\alpha_2\in[0,1]$, define
\[
\rho_{\bar\alpha}(Z\mid H_t,\mu)=\lambda \text{CVaR}_{\alpha_1}(Z\mid H_t,\mu) +(1-\lambda)\text{CVaR}_{\alpha_2}(Z\mid H_t,\mu)
\]
and notice that $\rho_{\bar\alpha}(Z\mid H_t,\mu)=\lambda\mathbb E\left(Z\mid H_t,\mu\right) +(1-\lambda) \text{CVaR}_\alpha\left(Z\mid H_t,\mu\right)$ when the vector of CVaR confidence intervals is given by $\bar\alpha=(1,\alpha)$.
\begin{theorem}\label{thm:decompose_2CVaR}
For any $t\geq 0$, denote by $Z \doteq (Z_{t+1},Z_{t+2},\dots)$ the cost sequence from time $t+1$ onwards. The conditional mean-CVaR risk metric under policy $\mu$ obeys the following decomposition:
\[
 \rho_{\bar\alpha}(Z\mid H_t,\mu)= \max_{\xi\in \U_{\text{2CVaR}}(\bar\alpha, P(\cdot|x_t,a_t))}\mathbb E[\mathbf{S}_\lambda(\xi(x_{t+1}))\cdot \rho_{\bar\alpha\mathbf{S}_\lambda(\xi(x_{t+1}))}(Z\mid H_{t+1})\mid H_t]
\]
where $\bar\alpha=(\alpha_1,\alpha_2)$  is the vector of CVaR confidence intervals. The risk envelop is given by
 \[
\U_{\text{2CVaR}}(\bar\alpha,P(\cdot|x_t,a_t))\!=\!\left\{\xi=(\xi_1,\xi_2):\xi_i(x_{t+1})\!\in\!\bigg[0,\frac{1}{\alpha_i}\bigg],\sum_{x_{t+1}\in\mathcal X}\xi_i(x_{t+1})P(x_{t+1}|x_t,a_t)=1,\,\forall i\right\},
\]
$\mathbf{S}_\lambda(\xi):\reals^2\mapsto\reals$ is a linear operator given by $\lambda\xi_1+(1-\lambda)\xi_2$ and $a_t$ is the control input induced by policy $\mu_t(h_t)$.
\end{theorem}
Now we extend the above analysis to Bellman recursion. With the generic state space $\mathcal Y=[0,1]^2$, we now define the optimal Bellman operator at any $(x,y)\in\mathcal X\times\mathcal Y$,
\begin{equation}\label{bellman_2CVaR}
\mathbf T[V](x,y)= \min_{a\in\mathcal A}\left[C(x,a)+\gamma\max_{\xi\in \U_{\text{2CVaR}}(y, P(\cdot|x,a))}\sum_{x'\in\mathcal X}\mathbf{S}_\lambda(\xi(x'))V\left(x',y\mathbf{S}_\lambda(\xi(x'))\right)P(x'|x,a)\right].
\end{equation}
Based on the decomposition result from Theorem \ref{thm:decompose_2CVaR},  we now have the result on the convergence of Bellman recursion, analogous to Theorem \ref{thm:MDP_CVaR} and \ref{thm:policy}, showing that the fixed point solution of $\mathbf T[V](x,y)=V(x,y)$ is unique and equals to the solution of \eqref{eq:mean-CVaR} with $x_0=x$ and $y_0=(1,\alpha)$.
\begin{theorem}\label{thm:MDP_2CVaR}
For any state $x\in\mathcal X$ and $y=(y_1,y_2)\in[0,1]^2$, the fixed point solution of $\mathbf T[V](x,y)=V(x,y)$ is unique and is equal to $V(x,y):=\min_{\mu\in\Pi_H} \lambda \text{CVaR}_{y_1}\left(\lim_{T\rightarrow\infty}\mathcal C_{0,T}\mid x_0,\mu\right) +(1-\lambda) \text{CVaR}_{y_2}\left(\lim_{T\rightarrow\infty}\mathcal C_{0,T}\mid x_0,\mu\right)$. Furthermore, let $\mu^*=\{\mu_0,\mu_1,\ldots\}\in\Pi_H$ be a policy recursively defined as in \eqref{eq:policy_construct} with two-dimensional augmented state $\{y_j\}$ and initial condition $y_0=(1,\alpha)$. Then $\mu^*$ is an optimal policy for the mean-CVaR problem \eqref{eq:mean-CVaR} with initial condition $x_0$ and CVaR confidence level $\alpha$.
\end{theorem}

Extending the interpolation-based CVaR value iteration (Algorithm \ref{alg:CVI}) for this case is straightforward, using a 2-D linear interpolation for $yV(x,y)$.
\end{document}